\documentclass[journal]{IEEEtran}
\usepackage[utf8]{inputenc}
\usepackage{graphicx}
\usepackage{subcaption}
\usepackage{cite}
\usepackage{amsmath,amssymb,multirow}
\usepackage{amsmath,amssymb}
\DeclareMathOperator{\E}{\mathbb{E}}
\newcommand{\squeezeup}{\vspace{-2.5mm}}
\DeclareMathOperator*{\argmax}{arg\,max}

\usepackage{algorithm,algorithmic,float,balance,cite,tabularx,booktabs}

 \usepackage{scalerel}
 \usepackage{tikz}
 \usetikzlibrary{svg.path}
 \usetikzlibrary{svg.path}

 \definecolor{orcidlogocol}{HTML}{A6CE39}
 \tikzset{
   orcidlogo/.pic={
     \fill[orcidlogocol] svg{M256,128c0,70.7-57.3,128-128,128C57.3,256,0,198.7,0,128C0,57.3,57.3,0,128,0C198.7,0,256,57.3,256,128z};
     \fill[white] svg{M86.3,186.2H70.9V79.1h15.4v48.4V186.2z}
                  svg{M108.9,79.1h41.6c39.6,0,57,28.3,57,53.6c0,27.5-21.5,53.6-56.8,53.6h-41.8V79.1z M124.3,172.4h24.5c34.9,0,42.9-26.5,42.9-39.7c0-21.5-13.7-39.7-43.7-39.7h-23.7V172.4z}
                  svg{M88.7,56.8c0,5.5-4.5,10.1-10.1,10.1c-5.6,0-10.1-4.6-10.1-10.1c0-5.6,4.5-10.1,10.1-10.1C84.2,46.7,88.7,51.3,88.7,56.8z};
   }
 }
 \newcommand\orcidicon[1]{\href{https://orcid.org/#1}{\mbox{\scalerel*{
 \begin{tikzpicture}[yscale=-1,transform shape]
 \pic{orcidlogo};
 \end{tikzpicture}
 }{|}}}}

\usepackage{amsmath,amssymb}
\usepackage{hyperref}
\usepackage{tablefootnote}

\usepackage{graphicx}
\usepackage{textcomp}
\usepackage{xcolor}
\usepackage{color,soul}
\usepackage{dirtytalk}

\usepackage{hyperref}
\usepackage{lipsum}

\begin{document}

\title{Improving ClusterGAN Using Self-Augmented Information Maximization of Disentangling Latent Spaces}

 \author{\IEEEauthorblockN{Tanmoy Dam~$^{\orcidicon{0000-0003-3022-0971}}$,
  Sreenatha G. Anavatti~$^{\orcidicon{; 0000-0002-4754-8191}}$,
  Hussein A. Abbass,~\IEEEmembership{Fellow,~IEEE}~$^{\orcidicon{0000-0002-8837-0748}}$}\\
  \IEEEauthorblockA{School of Engineering and Information Technology, University of New South Wales Canberra,  Australia.}}

\maketitle

\begin{abstract}

Since their introduction in the last few years, conditional generative models have seen remarkable achievements. However, they often need the use of large amounts of labelled information. By using unsupervised conditional generation in conjunction with a   clustering inference network, ClusterGAN has recently been able to achieve impressive clustering results. Since the real conditional distribution of data is ignored, the  clustering inference network can only achieve inferior clustering performance by considering only uniform prior based generative samples. However, the true distribution is not necessarily balanced. Consequently, ClusterGAN fails to produce all modes, which results in sub-optimal clustering inference network performance. So, it is important to learn the prior, which tries to match the real distribution in an unsupervised way. In this paper, we propose self-augmentation information maximization improved ClusterGAN (SIMI-ClusterGAN) to learn the distinctive priors from the data directly. The proposed SIMI-ClusterGAN consists of four deep neural networks: self-augmentation prior network, generator, discriminator and clustering inference network. The proposed method has been validated using seven benchmark data sets and has shown improved performance over state-of-the art methods. To demonstrate the superiority of SIMI-ClusterGAN performance on imbalanced dataset, we have discussed two imbalanced conditions on MNIST datasets with one-class imbalance and three classes imbalanced cases. The results highlight the advantages of SIMI-ClusterGAN.


\end{abstract}

\section{introduction}

Clustering is a popular unsupervised representation learning method, which has been widely studied in computer vision and machine learning communities such as image segmentation \cite{chuang2006fuzzy}, visual features representation learning \cite{caron2018deep}, and 3D object recognition \cite{vincent2010stacked}. The lower dimensional representation of high dimensional semantics data has been described by many latent-space based clustering methods, such as DCN \cite{yang2017towards}, DEC \cite{xie2016unsupervised}, Dual-AE \cite{yang2019deep}, and ClusterGAN \cite{mukherjee2019clustergan}. 

Latent space clustering methods have been developed based upon the clustering-assignment objective in the latent-space. Therefore, the objective still lies in the discriminative learning representation blending with data reconstruction loss. This discriminative lower dimensional representation can capture all discrete factors in the lower dimensional latent space  
such as data variations within the same group, which are the key factors for clustering objective. However, it fails to reconstruct the  real data due to the clustering assignment objective. Hence, similarities based clustering methods (such as K-means) have been incorporated with latent space to determine the clustering assignments~\cite{yang2017towards}. Further improvement using latent-space clustering objectives and pretrained reconstruction are employed to determine the optimal lower cluster assignment representation~\cite{min2018survey}. However, this is a very laborious task to obtain both optimal lower representation and a reconstruction of the real data at the same time. To obtain this representation, the high-dimensional data is first represented in lower dimensional latent space and then different distance-metric clustering objectives are applied on latent space to determine clustering assignments~\cite{xie2016unsupervised} \cite{macqueen1967some}.

Generative adversarial networks (GAN) are one of the most effective tools to learn an implicit function~\cite{goodfellow2014generative, ma2020snegan, he2021finger,li2019af} from complex real data distribution~\cite{isola2017image, vincent2010stacked, oord2016wavenet, tang2020multi, jing2020self, liu2020cgan, mukherjee2019clustergan, zheng2021reward}. The learning process is formulated as an adversarial game using the min-max principle. The generator network takes the sample from the noisy latent distribution to estimate the real distribution, while the discriminator network is trained to discriminate the real from fake samples, respectively. For learning complicated data distributions, generative models have been extensively studied in most circumstances. When it comes to downstream tasks like clustering, 
the estimation of the posterior distribution of latent inference from the data is an intractable problem in GAN models. Therefore, the researchers developed a mode-matching network for estimating downstream tasks \cite{mukherjee2019clustergan, chen2016infogan}. In a mode matching network, a clustering inference network estimates the multi-modal latent prior through bounding loss. Thus, an inference network works as a regularisation of the GAN objective where each mode is defined for each class of real data distribution. However, it is well-defined that mode matching network  can't be fully enforced the clustering objective on the generative samples when dealing with high-dimensional data \cite{arora2018gans, karras2018progressive,mishra2020effect} . Therefore, for high-dimensional data like CIFAR-10, STL-10, the mode matching network \cite{mukherjee2019clustergan, chen2016infogan} is dependent on the semantics of the data, which may not be directly related to the labels(class) information. In addition, in real-life scenarios, the classes of real distribution are not uniformly distributed. For example, the MNIST dataset contains ten distinctive classes, where mode matching methods try to estimate ten distinctive modes through ten uniform priors \cite{mukherjee2019clustergan, chen2016infogan}. Although the latent priors originate from ten different distributions, the uniform prior-based generative model can only create major classes. Thus, the inverting inference network fails to provide appropriate clustering performance. As a solution, we need to include regularisation that compel inference networks to accurately estimate downstream tasks and create all unique modes in imbalanced situations. To tackle both the issues, we propose a self augmented information maximization improved ClusterGAN (SIMI-ClusterGAN) in this paper. SIMI-ClusterGAN has four deep neural networks: self-augmentation prior network, generator, discriminator and clustering inference encoder. Prior network is used to learn the discrete representation of data in discrete space by self-augmentation maximization principle. This learned prior blends with continuous variable to form a discrete noise latent space. Then, generative network takes continuous-discrete Gaussian noise latent space prior to map in real data domain. Clustering inference encoder predicts the lower-level representation of the generated as well as real data through clustering assignment indices. To enforce the  disentanglement clustering representation of real data, the weight sharing generator-encoder networks pair have been utilised. We propose three additional loss functions to modify the ClusterGAN objectives. These three losses are used to create distinctive latent-space to help maintaining diversity in generative samples in addition to improving the clustering performance.

In summary, the main contributions of this work are as follows:

\begin{itemize}
  \item  We propose self-augmented mutual information maximization based disentangle categorical priors (one-hot encode)learning from the data.
  
  \item The learned discrete categorical priors with continuous Gaussian noise are utilised to form smooth disentangle noise latent spaces. This learned mixture of continuous-discrete disentangle latent spaces is used to generate the real class distribution of the data. 
  
  \item Three additional loss functions are introduced to modify ClusterGAN objectives for improved performance. Theses loss functions account for reconstruction, prior bounding and cross modality.
  
  \item A comparison of the the proposed SIMI-ClusterGAN and other GAN clustering methods is presented to validate the superiority of the proposed method in clustering tasks. We have used seven datasets including CIFAR-10 and STL-10 to accommodate complex high-dimensional data. We further demonstrate the superiority of our proposed method for imbalanced dataset by comparing the performance with ClusterGAN under two different imbalanced conditions.  
  
\end{itemize}

\section{Related Work}

\subsection{Disentangle latent space clustering} 

Most of the latent-space based approaches in practice have the encoder-decoder (Enc-Dec) structure where lower dimensional encoding latent code is used for clustering tasks~\cite{chang2017deep,  yang2016joint,  guo2017improved,xie2016unsupervised}. In deep embedding clustering (DEC)~\cite{xie2016unsupervised}, an Enc-Dec structure with reconstruction loss has been considered to train the model parameters. Once pre-training has been done, then a clustering network is added over the Enc network for further training with clustering similarities loss. In improved DEC~\cite{guo2017improved}, to maintain the local structure in the latent space, reconstruction loss is also considered in the final stage. The whole learning process is jointly trained as in DEC~\cite{xie2016unsupervised}. The Deep clustering network(DCN) is developed based on the Enc-Dec and K-means clustering objective~\cite{yang2016joint}. In order to maintain the local semantics in latent space, the reconstruction loss is adopted along with the K-means objective. However, the performance of the DCN method is still not seen to be adequate. 

The variation of disentangle factor in autoencoder based generative models can give explainable semantic latent code from the data. Most of the disentangle representations are associated with defining the separate latent spaces using variational factors~\cite{mathieu2016disentangling, gonzalez2018image,tschannen2018recent, pan2020loss} in which a single-stage or two-stage learning procedure can be found. Single stage disentanglement mostly deals with variations of two factors~\cite{mathieu2016disentangling, hadad2018two, patacchiola2019autoencoders,ye2020probabilistic} or three factors~\cite{ gonzalez2018image,tschannen2018recent}.  In all single-stage methods, the disentanglement has improved due to the partial knowledge of label information through cross-entropy loss. Meanwhile, two-stage disentanglement methods, such as $\beta$-TCVAE~\cite{chen2018isolating}, $\beta$-VAE~\cite{higgins2016beta}, factor-VAE~\cite{kim2018disentangling}, and joint continuous-discrete factors VAE~\cite{dupont2018learning} methods create separate latent spaces without knowledge of class-label information. These methods mostly deal with autoencoder structures with a Gaussian prior distribution in the encoded space. The disentanglement is accounted through reconstruction grade and regularisation of latent code factorization even though the real world data require more disentanglement\textquoteright s factors (eg. labels), which can\textquoteright t be obtained directly with continuous latent spaces. Consequently, joint VAE~\cite{dupont2018learning} came up with continuous-discrete factors of variations into the latent space.

\subsection{Generative prior-based inverse latent space clustering}

The multi-modal prior based latent space clustering can be segregated into two ways, 1) by applying a mixture  of continuous distributions, such as the GMM \cite{gurumurthy2017deligan}, and 2) by applying a combination of discrete and continuous distributions, such as the InfoGAN \cite{chen2016infogan} and ClusterGAN\cite{mukherjee2019clustergan}. The latter is the more popular of the two and is often accomplished by the concatenation of discrete and continuous random variables. In InfoGAN, the  disentanglement of latent factor is maximised between the latent code and the generated data. The modification of infoGAN, known as ClusterGAN \cite{mukherjee2019clustergan}, works on the same principle except for the bounded non-smooth latent space. The non-smooth discrete disentangle latent space is based on mixture of continuous-discrete (one-hot code) latent variables. The one-hot code discrete variable is used to measure the clustering performance by the encoder inference network. However, the continuous-discrete latent space variables are unable to complete disentanglement (e.g. categories) in ClusterGAN. It is reasonable to assume that the cluster characteristics of real data will not be replicated in the generated data, resulting in incorrect coverage of clusters in the generated data, as described in \cite{chen2016infogan}. To tackle such an issue,  more disentanglement in latent space is observed in ~\cite{mishra2020effect}. However, the proposed NEMGAN ~\cite{mishra2020effect}, that has the same ClusterGAN structure except a mode engineering network is used to learn the discrete prior from the data. The Mode network is updated by minimizing the KL divergence loss between mutual information data and mode network output. Thus, mode network output depends on mutual information of the data which is more relevant than the uniform discrete distribution of ClusterGAN. While updating the mode network, NEMGAN accounts for partial true label information. However, the performance is not significantly improved without partial information of true labels compared to the ClusterGAN. Therefore, NEMGAN is not a completely unsupervised method; instead, we would say it is a semi-supervised method.

In SIMI-ClusterGAN, we have proposed a two-stage approach to improve ClusterGAN performance. For learning the discrete prior assignments from the data, the self-augmented information maximization principle is employed. Once the learned prior from the data is obtained, it is used to form continuous-discrete mixtures to improve our SIMI-ClusterGAN performance. Unlike NEMGAN, the clustering inference network is used to separate the continuous-discrete variable without any supervision.

\section{Proposed approach}
\begin{figure*}
\includegraphics[scale=0.5]{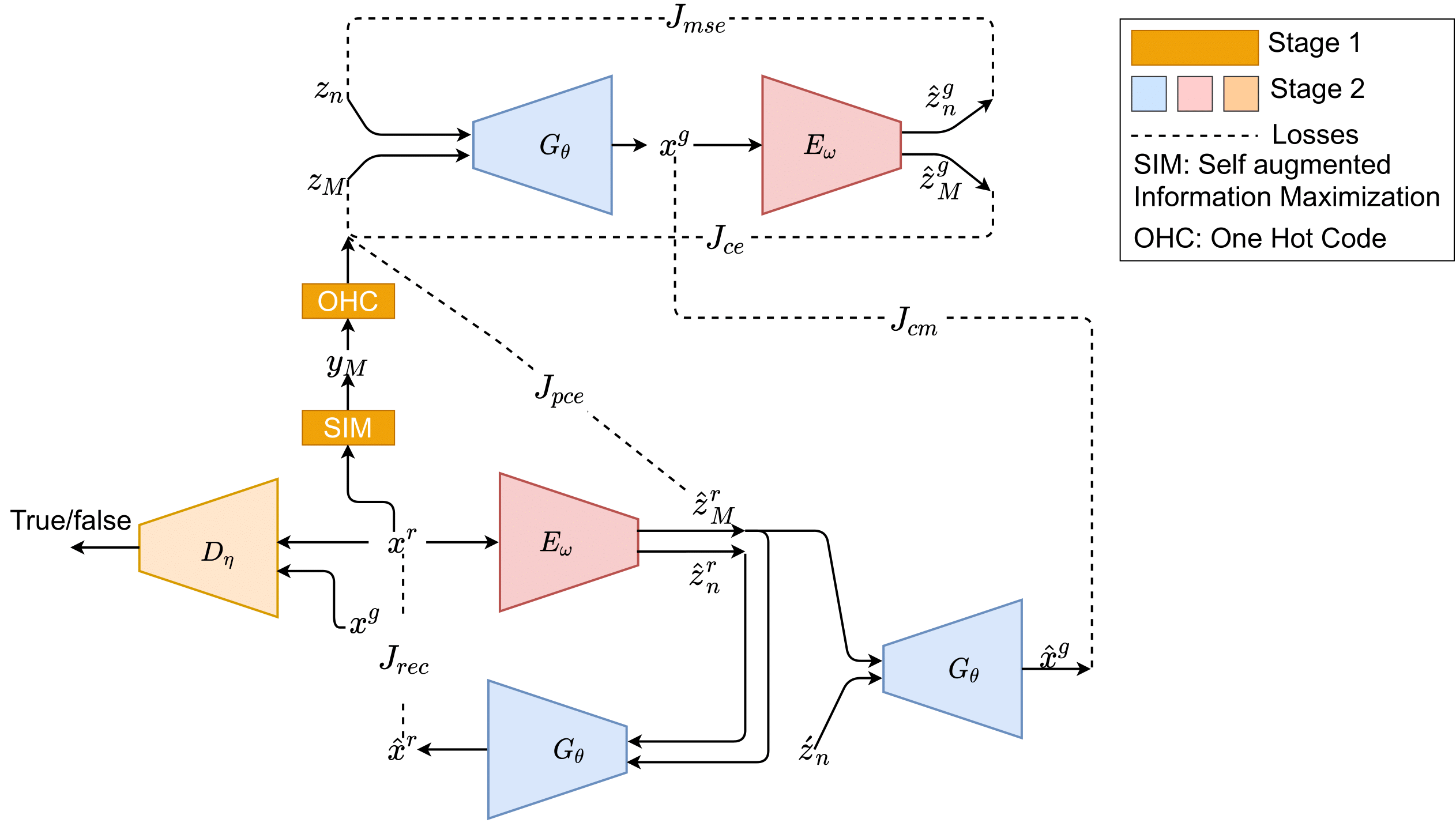}
\caption{The SIMI-ClusterGAN Architecture. $E_{\omega}$ and $G_{\theta}$ are sharing networks.}
\label{fig:SAIM-ClusterGAN architecture}
\end{figure*}

This section begins with a brief introduction to the ClusterGAN algorithm. The discrete prior is then discovered by applying the SIM principle directly to the data. We created our novel SIMI-ClusterGAN approach by combining this learnt discrete prior with three additional losses. The  architecture of the proposed method is depicted in Figure \ref{fig:SAIM-ClusterGAN architecture}.

\subsection{ClusterGAN}

Let, the given $i$-th sample of real data  $\{x_i^r\}_{i=1}^N \in X \in p_r$, represent the $M$-discrete representable form by the function $Y=s(X)$, where $Y \in \{y_1,y_2,..., y_M\} \subseteq \textbf{c} $. Where, $\textbf{c} \in \{c_1,c_2,..., c_M\}$ is the ground truth classes. The ClusterGAN objective is estimate the $s(.)$ through adversarial GAN game principle.  Generative adversarial networks (GAN)~\cite{goodfellow2014generative} are defined by two neural network(NN) components, the generative component(G) and the discriminator component(D), which are parameterised by $\theta$ and $\eta$, respectively. The generator $(G_{\theta})$ takes noise distribution to map real distribution, $G_{\theta}: Z \mapsto X$ whereas the discriminator $(D_{\eta})$ is assigned a probabilistic value for the data sample of being real vs fake data, $D_{\eta}: X \mapsto R$. The GAN game is working on a two players min-max game principle defined as follows, 
\begin{equation}
    \mathop {\min }\limits_{\theta} \mathop {\max }\limits_{\eta}  \mathbb{E}_{x^r \sim p_r} [f(D_{\eta}(x^r))] +\mathbb{E}_{z \sim p_z}[f(1-D_{\eta}(G_{\theta}(z)))]
\end{equation}
where, the real sample $(x^r \in X)$ belongs to real probabilities distribution $x^r \in p_r$. $z \in p_z$ is the latent space samples, which are drawn from a known prior distribution ($p_z$)(e.g. multivariate Gaussian distribution). $p_g$ is the generated sample distribution which defines $x^g=G_{\theta}(z)$ and $f$ is the standard GAN type selection function. For vanilla GAN,  $f(x^r)= logx^r$ and for Wasserstein GAN, $f(x^r)= x^r$~\cite{gulrajani2017improved}. The adversarial GAN objective is to learn the function $f(.)$ $(p_r = f(p_z))$ through density estimation \cite{dam2021does}.

\subsubsection{Continuous-discrete prior based disentanglement in ClusterGAN}

In ClusterGAN, the multi-modal latent space($z \in Z$) is defined by a mixture of continuous and discrete prior distribution to form a discrete smooth manifold. Generally, a discrete smooth manifold mixture is obtained by cascading the normal distribution with uniform categorical distribution, and its disentanglement in latent space is controlled through categorical distributions. To be more specific, the latent space prior represents $z=(z_n, z_M)$ where $z_n=\mathcal{N}({\mu=0, \sigma^2*I_{d_n}})$ and $z_M \sim \mathcal{U}\{1,2, ...,M\}$, $\mathcal{U} \in R^M$ is the discrete uniform prior, which represents the number of $M$ classes present in the dataset. However, the variance$(\sigma)$ of the normal distribution ($\mathcal{N}$) is kept to a low value for the $G_{\theta}$ network so that all the discrete modes $(labels)$ can be generated with high purity. Hence, the sigma value~\cite{mukherjee2019clustergan} has been chosen to ($\sigma=0.10$) for all experiments in this paper. Thus, the continuous-discrete latent space ($z_n$) is bounded within $z \in(-0.6, 0.6)< 1$. This uniform discrete latent priors allow us to design an algorithm capable of detecting a number of clusters in the inverse generated space.

\subsubsection{Inference Clustering Network in ClusterGAN}

Many previous methods~\cite{creswell2018inverting,lipton2017precise} are mainly focused on the inverse latent space to reconstruct $(z^*)$ through an encoder inference $(E_{\omega})$ network. However, this bidirectional mapping can\textquoteright t guarantee the reconstruction of the latent space due to a lack of consistency between the samples and the latent space. Due to the non-convex optimization nature of the problem, the reconstruction always suffers from different latent space embedding $z$.  DeliGAN~\cite{gurumurthy2017deligan} has used different initialization to reconstruct the discrete latent space. Stochastic clipping of $z$ at each step is also found in~\cite{lipton2017precise}. However, none of the methods are related to the latent space based clustering.  

In ClusterGAN~\cite{mukherjee2019clustergan}, the bounded discrete latent space is computed by inference autoencoder network $(E_{\omega})$. The generator takes the mixture of continuous-discrete latent space $(z=z_n, z_M)$ to generate $x^g$ and the encoder estimates back to the latent space $(\hat z^g_n, \hat z^g_M) = E_{\omega}(x^g)$. The encoding cyclic loss for generated samples is computed by cross-entropy loss and reconstruction losses between $(z_n,z_M)$ and $(\hat z^g_n, \hat z^g_M)$. The generated samples cross entropy loss $(J_{ce})$ between $z_M$ and $\hat z^g_M$ is defined as follows,
\begin{equation}
    J_{ce}=  \E_{z \sim p_z}[l(z_M, E_{\omega}(G_{\theta}(z_n, z_M)))]
    \label{eq:CE_loss}
\end{equation}

where, $l$ is associated with the cross-entropy loss. The reconstruction loss $(J_{mse})$ between $z_n$ and $\hat z^g_n$ is defined by the mean square error (MSE) as follows,
\begin{equation}
    J_{mse}=  \E_{z \sim p_z} \parallel z_n- E_{\omega}(G_{\theta}(z_n, z_M))\parallel_2
    \label{eq:mse_loss}
\end{equation}

Finally, the three-player ClusGAN game directs attention to the following optimisation problem:
\begin{equation}
\begin{split}
    \mathop {\min}\limits_{\theta,\omega}  \mathop {\max }\limits_{\eta}  \E_{x \sim p_r} [f(D(x^r))] + \E_{z \sim p_z}[f(1-D(G(z)))] \\
    + \alpha_{cl} J_{ce} + \alpha_{mse} J_{mse}
\end{split}
\label{eq:ClusterGAN_equation}
\end{equation}

The regularisation coefficients ($\alpha_{cl}, \alpha_{mse}$) control the disentanglement factors in the encoded latent space. To determine the clustering accuracy, the real data distribution samples pass through the $E_{\omega}$ network to predict continuous and discrete variables. Afterwards, the $K$-means \cite{macqueen1967some} clustering algorithm is applied to concatenated  continuous and discrete vectors to obtain clustering performance. For downstream tasks like clustering, the cluster characteristics of real data can't be completely replicated through generated samples. This results in the disentanglement of encoded latent space, which may not be enough because the $E_{\omega}$ network has never been exposed to real distribution. As a result, adequate performance can't be obtained with this learning strategy. To tackle this issue, we first learn the lower level representation by using the SIM principle. Afterwards, the learned prior is used to guide more disentanglement in encoded latent space by using three additional losses. Unlike NEMGAN \cite{mishra2020effect}, both the learning stages don't consider any true level of information.
 
\subsection{Self-augmented information maximization(SIM) for discrete prior learning:}


We use SIM principle to learn the discrete prior(${y_1, y_2, ..., y_M}$) from the data directly. Our objective is estimate the $s(.)$ by maximised mutual information between  $X \& Y$.   According to Gomes et. al.~\cite{krause2010discriminative}, regularised information maximisation (RIM) maximizes the mutual information between the data and its discrete representation ($m=1,2,...,M$) by learning the prior network ($P_{\phi}(Y|X)$). Therefore, the regulariser based RIM objective function is defined as follows,
\begin{equation}
    R_{\phi}- \beta_{p}I(X:Y)
\label{eq:IMSAT_main}
\end{equation}
where, $I(.)$ measures the mutual information between the data and the discrete representation. The reguliser constraint, $R_{\phi}$, updates the prior network parameters $\phi$. $\beta_{p}$ controls the reguliser and mutual information. The prior network objective is to maximise the conditional probabilities based on the similar representation of the data~\cite{krause2010discriminative}. The independent conditional probabilities of given model, $P_\phi(y_1, y_2..., y_M|X)$, are represented in the following form,
\begin{equation} 
{P_{\phi}(y_1,y_2,...,y_M|X)= \prod_{m=1}^{M} P_{\phi}(y_m|X)}
\label{eq:IMSAT_conditional_probabilities}
\end{equation}
Maximising Eq.~\ref{eq:IMSAT_conditional_probabilities} is to represent the $M$ discrete classes based on the mutual information between data. 
Hu. et. al.~\cite{hu2017learning} suggested the regularisation $(R_{\phi})$  can be represented in a flexible formation, which is defined as a self-augmented training (SAT). The self-augmentation regularisation between data $x^r$ and its augmented variations $T(x^r)$ is represented as follows,
\begin{equation} 
{R_{SAT}({\phi; x^r, T(x^r))}= -\sum_{m=1}^{M} \sum_{y_m=0}^{M-1} P_{\hat \phi}(y_m|x^r)log P_{\phi}(y_m|T(x^r))}
\label{eq:IMSAT_RSAT_conditional_probabilities}
\end{equation}
where, $T(.)$ is a self augmentation function. $P_{\hat \phi}(y_m|x^r)$ is the current prediction of data $x^r$ and  $\hat \phi$ is the current update of $P$ network parameters. Equation~\ref{eq:IMSAT_RSAT_conditional_probabilities} represents the ability of the augmented data $(T(x^r))$ to push closer to the conditional probabilities $P_{\phi}(y_M|x^r)$.  Generally, for image data, the augmentation is mostly dealt with an affine transformation such as random rotation, scaling and shearing ~\cite{hu2017learning}. 

The local perturbation of data can't alter the invariant  nature and the local perturbation based self-augmentation can be defined as follows, 
\begin{equation}
    T(x^r) = x^r+ l_p 
\end{equation}
where $l_p$ is the local perturbation which does not change the data representation in low dimensional regions/manifold ~\cite{grandvalet2005semi}. The local perturbation regularisation method is based on the virtual adversarial training method~\cite{miyato2015distributional} in which the local perturbation is defined as follows,
\begin{equation}
l_p = \argmax_{\hat l_p} [R_{SAT}({\hat \phi}; x^r, x^r+l_p), \,\,\, \parallel{l_p}\parallel_{2} \leq \beta_t]
\label{eq:IMSAT_local_pert}
\end{equation}
The local perturbation solution of Eq.~\ref{eq:IMSAT_local_pert} can be easily obtained from~\cite{miyato2015distributional}. 

The discrete cluster representation of Eq.~\ref{eq:IMSAT_main} can be reviewed in-terms of the difference in the mutual information between the entropy and the marginal conditional entropy~\cite{krause2010discriminative} as follows,
\begin{equation}
R_{SAT}- \beta_{p}[H(Y)-\beta_{mu}H(Y|X)]
\label{eq:RSAT_mutual_loss}
\end{equation}
where, $H(Y)$ and $H(Y|X)$ represent the marginal entropy and conditionals entropy, respectively, which are calculated as follows,
\begin{equation}
\begin{split}
H(Y)= h(P_{\phi}(y_M))=h(\frac{1}{N} \sum_{n=1}^{N} P_{\phi}(y_M|x^r_n) )\\
H(Y|X)= \frac{1}{N} \sum_{n=1}^{N} h(P_{\phi}(y_M|x^r_n))
\end{split}
\label{eq:condtional_and_mutual_information}
\end{equation}
where, $h(.)$ is an entropy function. The marginal entropy and conditionals entropy are working together to represent the discrete representation ($y_M$) of the data. Hence, Eq.~\ref{eq:RSAT_mutual_loss} is the final optimization problem that can give us discrete prior index ($y_M$) from the data. After obtaining the learned prior, it has to be represented in one-hot-code categorical priors($z_M$) in the proposed SIMI-ClusterGAN. The one-hot code (OHC) representation of $y_M$ is defined by the following form,
 
\begin{equation}
z_M=1_M(y_M) = 
\begin{cases}
            $1$  & \text{when , $y_M \in M$ } \\
            $0$  & \text{when, $y_M \not \in M$} 
\end{cases}
\label{eq:one_hot_code_representation}
\end{equation}

\subsection{Inference network disentanglement representation in SIMI-ClusterGAN}

The generated samples ($x^g$) are only considered in the $(E_{\omega})$, to obtain cluster representations of the data in ClusterGAN. However, this can\textquoteright t guarantee that representation of data always lies in the categories of the labels~\cite{mukherjee2019clustergan}. Hence, a suitable disentanglement is required to enforce the representation to remain at categories labels. At the same time, it also handles the mode diversity issue. The $G_{\theta}$ takes the continuous $(z_n)$-discrete$(z_M)$ mixtures prior noise to generate samples $x^g$ and then, the generated samples passing through the $E_{\omega}$ to estimate the $\hat z^g_n$ and $\hat z^g_M$ by the two equations~\ref{eq:mse_loss} and~\ref{eq:CE_loss}, respectively. These two equations are known as cyclic loss in ClusterGAN objective Eq.~\ref{eq:ClusterGAN_equation}. To estimate the posterior distribution from the real data $(x^r)$, the $E_{\omega}$ encodes as ($\hat z^r_n$, $\hat z^r_M$). The $G_{\theta}$ takes the estimated categorical posterior $(\hat z^r_M)$ with varying continuous distribution, $(z_n)$, to generate realistic samples, $(\hat x^r)$. The encoding and decoding distributions match can be achieved by two networks $E_{\omega}$ and $G_{\theta}$, respectively~\cite{zhao2018adversarially}. This $E_{\omega}-G_{\theta}$ reconstruction loss is defined as follows,
\begin{equation}
    J_{rec}=  \mathbb{E}_{x^r \sim p_r} \parallel x^r- (G_{\theta}(E_{\omega}(x^r)))\parallel_2
    \label{eq:reconstruction_loss}
\end{equation}
This loss can be looked upon as the real-reconstructing loss between $E_{\omega}$ and $G_{\theta}$ pairs to force reconstruction levels disentanglement.   

The above cyclic losses and reconstruction losses are used to maintain the local level disentanglement between the real space, $(x^r)$, and the generated space, $(x^g)$. To enforce more disentanglement on the $E_{\omega}$, we have added one regularisation penalty on the bounded real discrete prior loss. We have used the same $E_{\omega}$ networks with sharing weights that will help to update the parameters on different constraints. To determine prior bounding loss, we draw samples from the real distribution, $x^r\in p_r$ and $E_{\omega}$ maps real samples into real disentangle representation variables, $\hat z^r_n$ and $\hat z^r_M$. The cross-entropy prior bounded loss is defined as follows,
\begin{equation}
    J_{pce}=  \mathbb{E}_{x^r \sim p_r}[l(z_M, E_{\omega}(x^r)]
    \label{eq:prior_bounded_loss}
\end{equation}
Moreover, to generate diverse samples within the same prior discrete representation ($z_M$)~\cite{mao2019mode}, we have also introduced cross-modality loss by considering two generated samples from the $G_{\theta}$ network. To generate two samples $(x^g ,  \hat x^g)$ from $G_{\theta}$, two different variants latent codes have been used with the same prior ($z_M$). They are obtained by varying the continuous variable ($z_n$) only while maintaining $z_M$ fixed. The latent variables for cross-modality is represented as $(z_1, z_2)$, where $z_1 = (z_n, z_M)$ and $z_2 = (z^{\prime}_n, \hat z^r_M = E_{\omega}(x^r))$. Cross-modality loss continues to generate diverse images and penalises the generator. Cross-modality is defined as follows,
\begin{equation}
    J_{cm}=  \mathbb{E}_{x^r \sim p_r} \mathbb{E}_{z \sim p_z} \parallel x_g- (G_{\theta}(z^{\prime}_n,E_{\omega}(x^r)))\parallel_2
    \label{eq:cross_modality_loss}
\end{equation}

\subsection{Final objective function of SIMI-ClusterGAN}

The proposed SIMI-ClusterGAN objective function incorporates all of the above defined losses along with ClusterGAN cyclic losses. Thus, the SIMI-ClusterGAN objective function is defined as follows,
\begin{equation}
\begin{split}
    \mathop {\min}\limits_{\theta,\omega}  \mathop {\max }\limits_{\eta} \E_{x^r \sim p_r} [f(D_{\eta}(x))] + \E_{z \sim p_z}[f(1-D_{\eta}(G_{\theta}(z)))] \\
    + \alpha_{cl} J_{ce} + \alpha_{mse} J_{mse} + \alpha_{re} J_{rec} + \alpha_{pcl} J_{pce} + \alpha_{cm} J_{cm}
\end{split}
\label{eq:SIMI-ClusterGAN_equation}
\end{equation}
where, the regularisation coefficients($\alpha_{re},\alpha_{pcl},\alpha_{cm} $) are used to maintain disentanglement of encoded latent space and diversify the generated samples.

\begin{algorithm}
\caption{SIMI-ClusterGAN algorithm}
\label{Algorithm:SIMI-ClusterGAN}
\begin{algorithmic}[1]
\REQUIRE {training data($D^{tra}$)= ${\left\{ {{{x^r} \in R^{d}},{y \in R^{M} }} \right\}^S}$,
testing data($D^{tst}$)=${\left\{ {{{x^r} \in R^{d}},{y \in R^{M} }} \right\}^T}$,
$epoch_1$=$60$, $epoch_2$=$200$, $B_p$=$256$, $B_m$=$30$, $\beta_{p} =0.1$, $\beta_{t}=0.25$, $\beta_{mu}= 4$, ADAM optimizer($\beta_1 =0.5$ \& $\beta_2=0.999$), $l_{r1}$= $0.002$ for $P_{\phi}$, $l_{r2}$= $0.0001$ for $D_{\eta}$, $G_{\theta}$, $E_{\omega}$, Critic iter($C_{iter}$)=1,  $\lambda$ =10, regularisation factors $\alpha_{cl}=10, \alpha_{mse}=10, \alpha_{re}=1, \alpha_{pcl}=10$ and  $\alpha_{cm}=1$ }

\ENSURE { $P_{\phi}$, $G_{\theta}$, $D_{\eta}$, $E_{\omega}$}


\STATE Phase 1: The Prior Learning Algorithm ($P_{\phi}$)
\FOR{$epoch\,\, \,\,in\,\,\,\, epoch_1$}
    
       \STATE Samples $x^r \in p_r$ drawn in a mini-batch $B_p$.
       \STATE Calculate the self-augmentation distance by using equ.(\ref{eq:IMSAT_local_pert}) 
       \STATE Maximize the discrete representation($y_M$) by optimizing the equ.(\ref{eq:RSAT_mutual_loss}) 
   
\ENDFOR

\end{algorithmic}
\begin{algorithmic}[1]
\STATE Phase 2: SIMI-ClusterGAN using the learned prior($P_{\phi}$)
\FOR{$epoch \, \, \, in \, \, \, epoch_2$}
   \FOR{iter  in  $C_{iter}$}
       \STATE Samples $x^r \in p_r$ drawn in a mini-batch $B_m$.
       \STATE Calculate discrete prior assignment $y_M$ = $P_{\phi}(x^r)$. 
       \STATE One hot code representation of learned prior, $z_M$ by using equation (\ref{eq:one_hot_code_representation}). 
       \STATE Samples $z_n \in p_z$ drawn in a mini-batch $B_m$. 
       \STATE the continuous-discrete mixture of latent space variable  $z= (z_n, z_M)$.
       \STATE $G_{\theta}(z)\rightarrow x^g $
       \STATE WGAN 1-GP stability applied on $D_{\eta}$ parameters. Discriminator parameters ($\eta$) are updated by the following equation, 
       $D_{\eta} \rightarrow  D_{\eta}(x^r)-D_{\eta}(x^g) + \lambda(||\nabla D_{\eta}(\hat x^r)||-0)^2$
       where, $\hat x^r = \alpha x^r+ (1-\alpha)x^g $
       
    \ENDFOR
    
\STATE Encodes the generated samples $E_{\omega}(x^g) \rightarrow (\hat z^g_n, \hat z^g_M)$.
\STATE Calculate the cyclic loss by using equations(\ref{eq:mse_loss}) and (\ref{eq:CE_loss}).
\STATE The reconstruction loss($J_{rec}$) between $E_{\omega}-G_{\theta}$ networks pair is obtained by equation (\ref{eq:reconstruction_loss}).
\STATE The prior bounded cross entropy loss ($J_{pce}$) is obtained by equation \ref{eq:prior_bounded_loss}.
\STATE The cross-modality loss ($J_{cm}$) is obtained by equation \ref{eq:cross_modality_loss}.
\STATE The generator($G_{\theta}$) network parameters ($\theta$)are updated by the following form,
$G_{\theta} \rightarrow  D_{\eta}(x^g)+ \alpha_{cl}J_{cl} + \alpha_{mse}J_{mse}+ \alpha_{cm} J_{cm} + \alpha_{re}J_{rec}$

\STATE The inference($E_{\omega}$) network parameters ($\omega$) are updated by the following form,
$E_{\omega} \rightarrow   \alpha_{cl} J_{cl} + \alpha_{mse}  J_{mse}+ \alpha_{cm}  J_{cm} + \alpha_{re}  J_{rec} + \alpha_{pcl}  J_{pce} $

\IF{$epoch_2 \mathbin{\%} 1== 0$}
    \STATE /* Encode latent spaces for $D^{tst}$ */
    \STATE  $E_{\omega}(x^r \in D^{tst} )\rightarrow (\hat{z}_n^r , \hat{z}_M^r)$
    \STATE Apply K-means on $(\hat{z}_n^r , \hat{z}_M^r)$ to calculate ACC and NMI values.
\ENDIF

\ENDFOR
\end{algorithmic}
\end{algorithm}

\section{Results \& Experiments}
We validate the proposed SIMI-ClusterGAN performance on several benchmark datasets. We used an imbalanced dataset under two imbalanced situations to verify the proposed method's superiority over ClusterGAN. We also examine the extensive ablation studies of each component of the proposed method\textquoteright s objective function to evaluate the clustering performance.

\subsection{Datasets}  
SIMI-ClusterGAN clustering performance has been evaluated on several benchmark datasets such as MNIST, Fashion-MNIST, USPS, Pendigits, $10\times{_-}73k$, STL-10 \cite{coates2011analysis} and CIFAR-10 \cite{krizhevsky2009learning} datasets. The information about these datasets and their corresponding latent space dimensions are provided in Table~\ref{data_description}. The SIMI-ClusterGAN algorithm is provided in Algorithm~\ref{Algorithm:SIMI-ClusterGAN}

\begin{table}[]
\caption{Data and latent variable Dimensions}
\begin{tabular}{l|l|l|l|l|l}
\textbf{Datasets}      & \textbf{Samples} & \textbf{Labels} & \textbf{Dimensions} $\in R^d$ & $z_n$ & $z_M$\\ \hline
MNIST \cite{xiao2017fashion}        & 70k     & 10    & $1 \times 28 \times 28$ & 30 & 10   \\
Fashion-MNIST \cite{xiao2017fashion} & 70k     & 10     & $1 \times 28 \times 28$  & 30 & 10   \\
USPS   \cite{yang2019deep}        & 9,298     & 10     & $1 \times 16 \times16$  & 30 & 10   \\
Pendigits  \cite{mukherjee2019clustergan}    & 10992   & 10     & $1 \times 16$  & 5 & 10     \\
$10  \times{\_}73k$ \cite{mukherjee2019clustergan}     & 73233   & 8      & $1 \times 720$      & 30 & 8 \\ \hline
CIFAR-10 \cite{hu2017learning}     & 60k     & 10     & $3 \times32 \times32$  & 50 & 10   \\
STL-10   \cite{hu2017learning}     & 13k    & 10     & $3 \times 96 \times 96$   & 100 & 10  
\\ \hline
\end{tabular}
\label{data_description}
\end{table}
 
\subsection{Simulations Implementation}
Four different neural networks $P_{\phi}, G_{\theta}, D_{\eta},  E_{\omega}$ have been developed for handling the four different tasks; prior assignment learning, generator, discriminator and clustering inference network, respectively. The $P_{\phi}$ is used to learn the prior indices from the data using the self-augmentation maximization principle. This prior assignment is used as the discrete one-hot code variable to train the improved clustering GAN. The data is  normalised in the range of $[-1, 1]$ for all datasets except STL-10 $[0,1]$. The tangent-hyperbolic and sigmoid activation functions are used in the last layer of the $G_{\theta}$ network.  However, for learning the clustering prior from the data, we have used a $P_{\phi}$ network with three MLP hidden layers given by ($d-1200-1200-M$) for the MNIST, Fashion-MNIST,  CIFAR-10, STL-10 datasets in which $d$ $\&$ $M$ represent input data dimension and cluster assignments index respectively. For lower dimensional data such as Pendigits, USPS, and $10\times{_-}73k$ , we have used the $P_{\phi}$ network as $d-256-256-M$. For MNIST and Fashion-MNIST, the conv-trans.conv layer with ReLU and LeakyReLU activation functions have been used in $G_{\theta}, D_{\eta}$ and  $E_{\omega}$ networks respectively \cite{mukherjee2019clustergan}. The clustering network $E_{\omega}$, has the same structure of $D_{\eta}$ to estimate continuous($\hat z_n$) and discrete ($\hat z_M$) variables. For Pendigits, $10\times{_-}73k$ and USPS datasets, we have used the two layers MLP models for $G_{\theta}, D_{\eta}$, and $E_{\omega}$.  Our proposed SIMI-ClusterGAN method has two learning stages. First, the discrete prior learns $M$ categorical discretization representations from data by using the $P_{\phi}$ network. The learning hyper-parameters are set to be $\beta_{p} =0.1$, mutual information coefficient $\beta_{mu} = 4$, perturbation coefficient $\beta_{t}= 0.25$ while updating the $P_{\phi}$ network. We have used the ADAM optimizer with learning rate of $0.002$.  Once the $M$ categorical representation is obtained then it is formulated into one-hot-code $(z_M) \in R^M$ to create $M$ discrete priors~\cite{mukherjee2019clustergan}. We have added the Gaussian Normal distribution, $z_n=\mathcal{N}({\mu=0, \sigma^2*I_{dn}}, \sigma = 0.1)$, into $z_M$ to create the $M$ discrete non-smooth surfaces. The noise samples are drawn from the continuous and discrete mixture variables to generate the realistic samples. Moreover, the learning rate of $G_{\theta}, D_{\eta}$, $E_{\omega}$ networks are set to $1e-4$. Similarly for $P_{\phi}$ network, we have adapted the ADAM optimiser with learning hyper-parameters $(\beta_1=0.5, \beta_2=0.999)$ \cite{mukherjee2019clustergan,dam2021does}. To determine the cluster labels from the real datasets, the $K-means$ ~\cite{macqueen1967some} algorithm is applied to estimate encode latent space from the $E_{\omega}$ network \cite{mukherjee2019clustergan}. The variation factors of each loss are controlled by five multiplication factors given by  $\alpha_{cl}=10$ \cite{mukherjee2019clustergan}, $\alpha_{mse}=10$ \cite{mukherjee2019clustergan}, $\alpha_{re}=1$, $\alpha_{pcl}=10$ and  $\alpha_{cm}=1$ . The simulations are implemented on PyTorch environment with NVIDIA GTX GPU.

\subsection{Clustering performance Evaluation} 
\begin{table*}[]
\caption{Quantitative analysis of the clustering performance}
\scalebox{1.2}{
\begin{tabular} {lllllllllll} \\ \hline
\multirow{ 2}{*}{Datasets}       & \multicolumn{2}{c}{MNIST}             & \multicolumn{2}{c}{Fashion-MNIST}     & \multicolumn{2}{c}{USPS}              & \multicolumn{2}{c}{Pendigits}         & \multicolumn{2}{c}{$10\times{_-}73k$}            \\
              & \multicolumn{2}{l}{Performance Index} & \multicolumn{2}{c}{Performance Index} & \multicolumn{2}{c}{Performance Index} & \multicolumn{2}{c}{Performance Index} & \multicolumn{2}{c}{Performance Index} \\ \hline
Models        & ACC               & NMI               & ACC               & NMI               & ACC                & NMI              & ACC               & NMI               & ACC                & NMI              \\ \hline
K-means  \cite{macqueen1967some}       & 0.532             & 0.500             & 0.474             & 0.501             & 0.668              & 0.601            & 0.793             & 0.730             & 0.623              & 0.577            \\
NMF \cite{lee1999learning}           & 0.471             & 0.452             & 0.500             & 0.510             & 0.652              & 0.693            & 0.670             & 0.580             & 0.710              & 0.690            \\
SC  \cite{shi2000normalized}          & 0.656             & 0.731             & 0.660             & 0.704             & 0.649              & 0.794            & 0.700             & 0.690             & 0.400              & 0.290            \\
AGGLO \cite{zhang2012graph}        & 0.640             & 0.650             & 0.550             & 0.570             & -                  & -                & 0.700             & 0.690             & 0.630              & 0.580            \\ \hline
DEC  \cite{xie2016unsupervised}         & 0.863             & 0.834             & 0.518             & 0.546             & 0.762              & 0.767            & -                 & -                 & -                  & -                \\
DCN   \cite{yang2017towards}          & 0.802             & 0.786             & 0.563             & 0.608             & -                  & -                & 0.720             & 0.690             & -                  & -                \\
JULE  \cite{yang2016joint}         & 0.964             & 0.913             & 0.563             & 0.608             & 0.950              & \textbf{0.913}            & -                 & -                 & -                  & -                \\
DEPICT  \cite{ghasedi2017deep}       & 0.965             & 0.917             & 0.392             & 0.392             & 0.899              & 0.906            & -                 & -                 & -                  & -                \\
SpectralNet \cite{shaham2018spectralnet}  & 0.800             & 0.814             & -                 & -                 & -                  & -                & -                 & -                 & -                  & -                \\
Dual-AE  \cite{yang2019deep}      & 0.978             & 0.941             & 0.662             & 0.645             & 0.869              & 0.857            & -                 & -                 & -                  & -                \\ \hline
InfoGAN   \cite{chang2017deep}     & 0.890             & 0.860             & 0.610             & 0.590             & -                  & -                & 0.720             & 0.730             & 0.620              & 0.580            \\
ClusterGAN  \cite{mukherjee2019clustergan}     & 0.950             & 0.890             & 0.630             & 0.640             & $0.798^*$            & $0.703^*$            & 0.770             & 0.730             & 0.810              & 0.730            \\
GAN with bp \cite{mukherjee2019clustergan}  & 0.950             & 0.900             & 0.560             & 0.530             & -                  & -                & 0.760             & 0.710             & 0.650              & 0.59             \\
CaTGAN  \cite{springenberg2015unsupervised}      & 0.890             & 0.900             & 0.550             & 0.600             & -                  & -                & -                 & -                 & -                  & -                \\
GANMM    \cite{yu2018mixture}       & 0.640             & 0.610             & 0.340             & 0.270             & -                  & -                & -                 & -                 & -                  & -                \\
NEMGAN\_v  \cite{ mishra2020effect}   & 0.960             & 0.910             & 0.650             & 0.610             & -                  & -                & -                 & -                 & -                  & -                \\ \hline
SIMI-ClusterGAN & \textbf{0.986}             & \textbf{0.958}            & \textbf{0.745}             & \textbf{0.705}             & \textbf{0.951}              & 0.894            & \textbf{0.857}             & \textbf{0.825}             & \textbf{0.917}              & \textbf{0.878}   \\ \hline  
\end{tabular}}
No available data : $(-)$, simulated :$*$ .
\label{Table:CLustering_performance_4datasets}
\end{table*}

We mainly focus on two popular evaluation indices given by accuracy (ACC)~\cite{xie2016unsupervised} and normalised mutual information (NMI)~\cite{xie2016unsupervised} for comparing the clustering performance. The ACC is defined as follows,
\begin{equation}
    ACC=max_{\delta}\frac{\sum_{i=1}^{N}\boldsymbol{1}\left\{ c_{i}=\delta(y^k_{i})\right\} }{N}
\end{equation}
where, $c_i \in R^M$ is the $i$-th ground truth labels, $y^k_i$ is $i$-th predicted labels which is obtained by applying $K$-means \cite{macqueen1967some} method on encoded latent-space($E_{\omega}$) \cite{mukherjee2019clustergan} . The $\delta$ covers all potential one-to-one mappings between true and predicted labels of data. This measure identifies the best match between an unsupervised algorithm's cluster prediction and a ground truth assignment. The Hungarian method efficiently computes the optimum mapping \cite{kuhn1955hungarian}. Similarly, NMI is defined as follows,
\begin{equation}
NMI=\frac{2\times I(Y;\textbf{c})}{H(Y)+H(\textbf{c})}
\end{equation}
where, $I(Y;\textbf{c})$ is defined mutual information between predicted and ground truth labels. NMI determines similarity between the true class and the predicted class labels, which is bounded by $0$ (worst similarity) to $1$ (ultimate similarity). 

Table~\ref{Table:CLustering_performance_4datasets} represents the clustering results of SIMI-ClusterGAN and other state-of-the art methods on the five benchmark datasets which are mostly dealt with single channel image datasets and a tabular dataset. The best evolution matrices are highlighted in the table~\ref{Table:CLustering_performance_4datasets} with 5 runs~\cite{mukherjee2019clustergan}. It is clearly seen from the Table~\ref{Table:CLustering_performance_4datasets} that deep neural network based clustering methods attained better performance compared to traditional machine-learning approaches. In comparison, our proposed SIMI-ClusterGAN achieved significant performance improvement for all datasets. For the MNIST dataset, our proposed method obtained better performance than the second best results of Dual-AE (ACC: .978 vs 0.986, NMI: .941 vs .958). While learning discrete prior ($P_{\phi}$) from the Fashion-MNIST dataset, we have used the following affine distortion to maximize the mutual information between categories($Y_M$). The following transformations are adopted to learn the $P_{\phi}$ network parameters,
\begin{itemize}
    \item Uniformly drawn samples along with random rotation by $\chi$ within $[-10^{\circ}, 10^{\circ}]$ 
    \item Uniformly drawn samples along with $x$ axis and $y$-axis by scaling factor $[.3, 1]$ 
    \item Uniformly drawn samples along with $x$ axis and $y$-axis by random shearing $(-0.3,0.3)$
\end{itemize}

The performance has improved significantly due to random affine transformation on the data since our proposed method is heavily dependent on the learning of the prior. Besides the affine transformations of data, self-augmentation is also applied to learn prior for the proposed SIMI-ClusterGAN method. The $ACC$ and $NMI$ values are $0.721$ and $0.675$, which are much better than the ClusterGAN. Similarly, for $10\times{_-}73k$ dataset, our proposed method performance is improved significantly compared with ClusterGAN (ACC: 0.810 vs 0.917 , NMI:  0.730 vs 0.878 ).  Specifically, for $16$-dimensional pendigits dataset, the K-means method outperformed all the GANs methods. But, our proposed SIMI-ClusterGAN obtained better performance on both evaluation indices (ACC:0.793 vs 0.857, NMI:0.730 vs 0.825). For the USPS dataset, the best NMI is obtained by JULE but our proposed method achieved the best results in ACC. The ClusterGAN method$^*$\footnote{\url{https://github.com/zhampel/clusterGAN}} obtained ACC and NMI values as 0.798 and 0.703 respectively.


\begin{figure}[ht!]
    \centering
    \begin{subfigure}[b]{0.23\textwidth}
        \includegraphics[width=\textwidth]{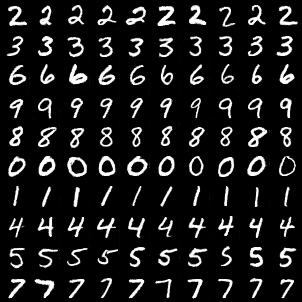}
        \caption{}
        \label{fig:mnist_all}
    \end{subfigure}
    ~ 
    \begin{subfigure}[b]{0.23\textwidth}
        \includegraphics[width=\textwidth]{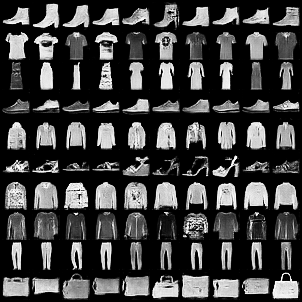}
        \caption{}
        \label{fig:fashion-mnist_all}
    \end{subfigure}
    \caption{Samples generated by SIMI-ClusterGAN for MNIST and Fashion-MNIST datasets. Each row represents each class. The generated classes maintained purity without mixtures of other classes}
    \label{fig:IClusterGAN_all_modes}
\end{figure}

\begin{figure}[ht!]
    \centering
    \begin{subfigure}[b]{0.23\textwidth}
        \includegraphics[width=\textwidth]{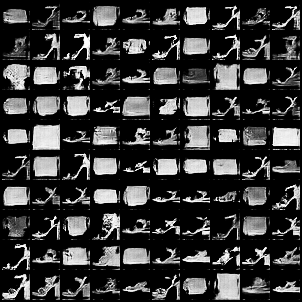}
        \caption{Sandal}
        \label{fig:mnist}
    \end{subfigure}
    ~ 
    \begin{subfigure}[b]{0.23\textwidth}
        \includegraphics[width=\textwidth]{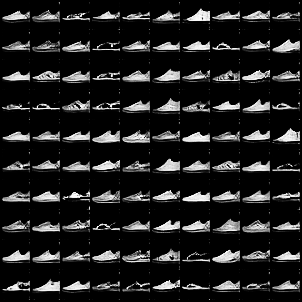}
        \caption{Sneaker}
        \label{fig:fashion-mnist}
    \end{subfigure}
    \caption{Samples generated by ClusterGAN for Fashion-MNIST dataset}
    \label{fig:clustergan_sandal_sneaker}
\end{figure}
\begin{figure}[ht!]
    \centering
    \begin{subfigure}[b]{0.22\textwidth}
        \includegraphics[width=\textwidth]{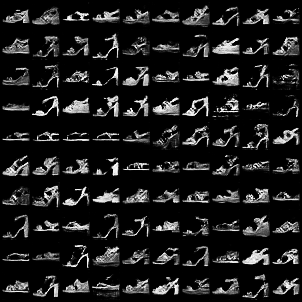}
        \caption{Sandal}
        \label{fig:mnist}
    \end{subfigure}
    ~ 
    \begin{subfigure}[b]{0.22\textwidth}
        \includegraphics[width=\textwidth]{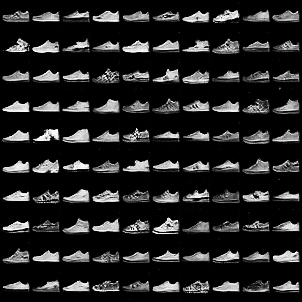}
        \caption{Sneaker}
        \label{fig:fashion-mnist}
    \end{subfigure}
    \caption{Samples generated by SIMI-ClusterGAN for Fashion-MNIST dataset}
    \label{fig:IClusterGAN_some_modes_fmnist}
\end{figure}

All the distinctive classes are fully achieved by SIMI-ClusterGAN method. It is observed from Figure~\ref{fig:IClusterGAN_all_modes} that the SIMI-ClusterGAN is able to generate all the modes without supervision with higher accuracy. For MNIST dataset, all the distinctive classes are represented in each row of Figure~\ref{fig:mnist_all}.Similarly, for Fashion-MNIST, our proposed method is able to generate all the classes without supervision as shown in Figure~\ref{fig:fashion-mnist_all}. For fair comparisons of our proposed method with ClusterGAN, we have plotted two distinctive generated classes("Sandal" and "Sneaker") for better understandings. ClusterGAN is not able to generate two distinctive classes without purity, as can be seen in Figure \ref{fig:clustergan_sandal_sneaker}. As shown in Figure ~\ref{fig:IClusterGAN_some_modes_fmnist},however, the proposed SIMI-ClusterGAN is capable of generating two different classes with a high degree of purity.

\subsection{Complex high-dimensional dataset performance}
We evaluate the proposed SIMI-ClusterGAN clustering performance on complex coloured image datasets: CIFAR-10 and STL-10. For learning the clustering prior, the colour image can't be directly used in the $P_{\phi}$ network because associated raw pixels of images are colour dominant~\cite{hu2017learning}. Therefore, for learning the prior clustering assignments, we have used 50-layer pretrained features extractor from raw data~\cite{hu2017learning, he2016deep}. We are not directly using the class information to $P_{\phi}$ network. Instead, the well-separated lower dimensional features are used to get the class assignment prior. For both the cases, the extracted features size is $2048$. Once, the prior assignment network is learned from extracted features data, the learned prior is used to train SIMI-ClusterGAN. The clustering performance of feature space data is listed in the Table~\ref{table:Complex_dataset_performance}. It is clearly observed from the table~\ref{table:Complex_dataset_performance} that our proposed method has outperformed other state-of-the-art methods. The uniform prior based ClusterGAN accuracy performance for CIFAR-10 is obtained as $15.9$. This is because the clustering loss in clustering inference network is focused on the generated samples' semantics instead of categories of data~\cite{mukherjee2019clustergan}. The ClusterGAN accuracy performance for STL-10 dataset is $12.6$. Similarly for the STL-10 dataset, our proposed method SIMI-ClusterGAN  obtained better results.

\begin{table}[]
\caption{Quantitative clustering performance for High-dimensional datasets}
\centering
\scalebox{0.8}{
\begin{tabular}{c|c|c|c|c}
\hline 
 & \multicolumn{2}{c|}{CIFAR-10} & \multicolumn{2}{c}{STL-10}\\
\hline 
Methods & ACC & NMI & ACC & NMI \\
\hline 
K-means\cite{hu2017learning} & 0.344 & - & 0.856 & - \\

DAE+KMeans\cite{hu2017learning} & 0.442 & - & 0.722 & - \\
 
DEC\cite{hu2017learning} & 0.469 & - & 0.781 & - \\
 
Linear RIM\cite{hu2017learning} & 0.403 & - & 0.735 & - \\
 
Deep RIM\cite{hu2017learning} & 0.403 & - & 0.925 & - \\
 
Linear IMSAT\cite{hu2017learning} & 0.407  & - & 0.917 & - \\

IMSAT(RPT)\cite{hu2017learning} & 0.455  & - & 0.928 & - \\

IMSAT(VAT)\cite{hu2017learning} & 0.456  & - & 0.941 & - \\
 
VADE \cite{jiang2017variational} & - & - & 0.844 & - \\
 
DDG \cite{yang2019deep} & - & - & 0.905 & - \\

Sarfaraz \cite{sarfraz2019efficient} & - & - & 0.952 & - \\
\hline 
ClusterGAN & $0.159^{*}$  & $0.025^{*}$ & $0.126^{*}$ & $0.096^{*}$ \\

SIMI-ClusterGAN & 0.512 & 0.412 & 0.954 & 0.861 \\
\hline 
Embedding feature-space clustering \cite{hu2017learning}, simulated :$*$ .
\end{tabular}}
\label{table:Complex_dataset_performance}
\end{table}

\begin{figure}[ht!]
    \centering
    \begin{subfigure}[b]{0.22\textwidth}
        \includegraphics[width=\textwidth]{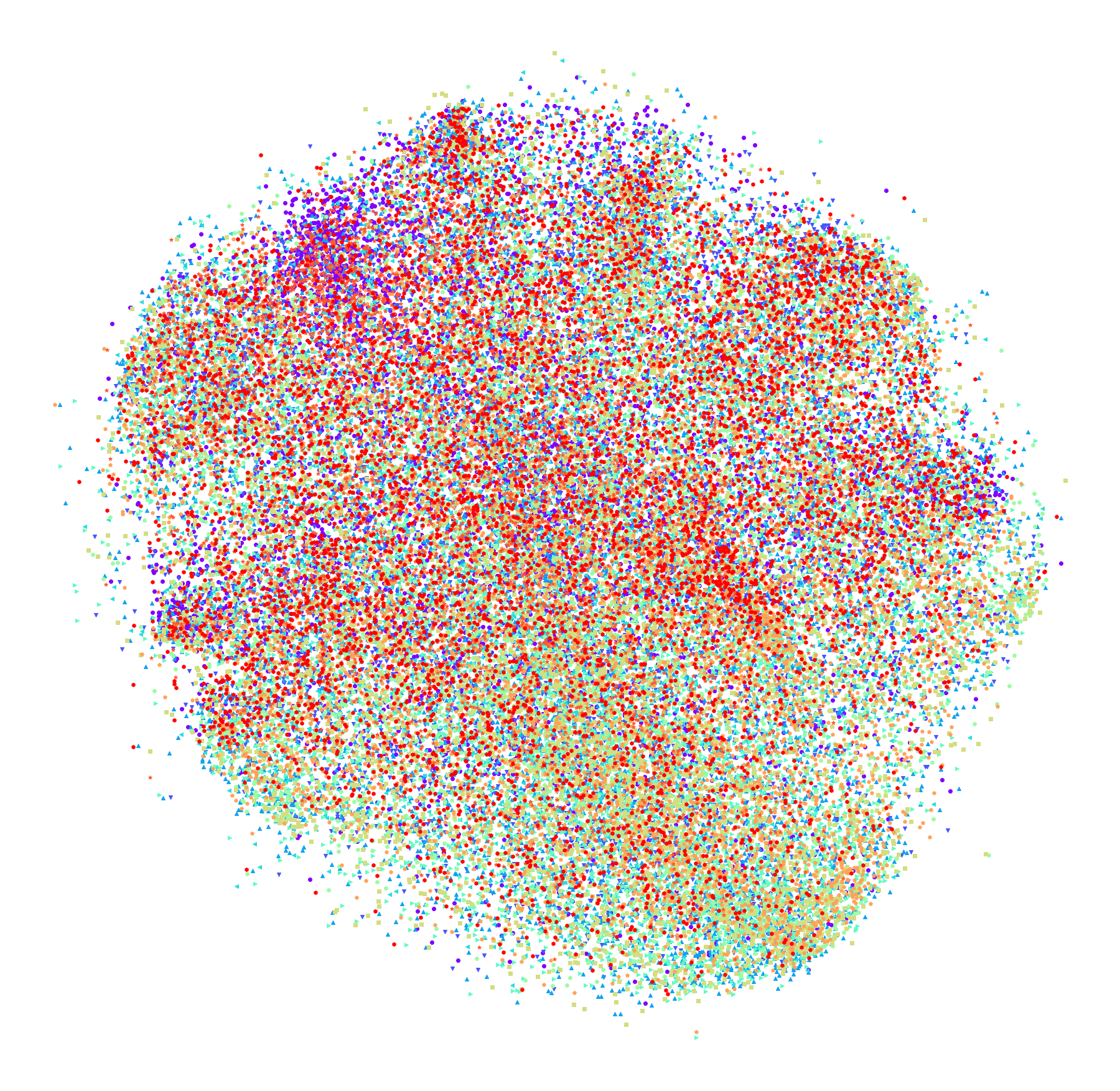}
        \caption{ClusterGAN}
        \label{fig:ClusterGAN_cifar10}
    \end{subfigure}
    ~ 
    \begin{subfigure}[b]{0.22\textwidth}
        \includegraphics[width=\textwidth]{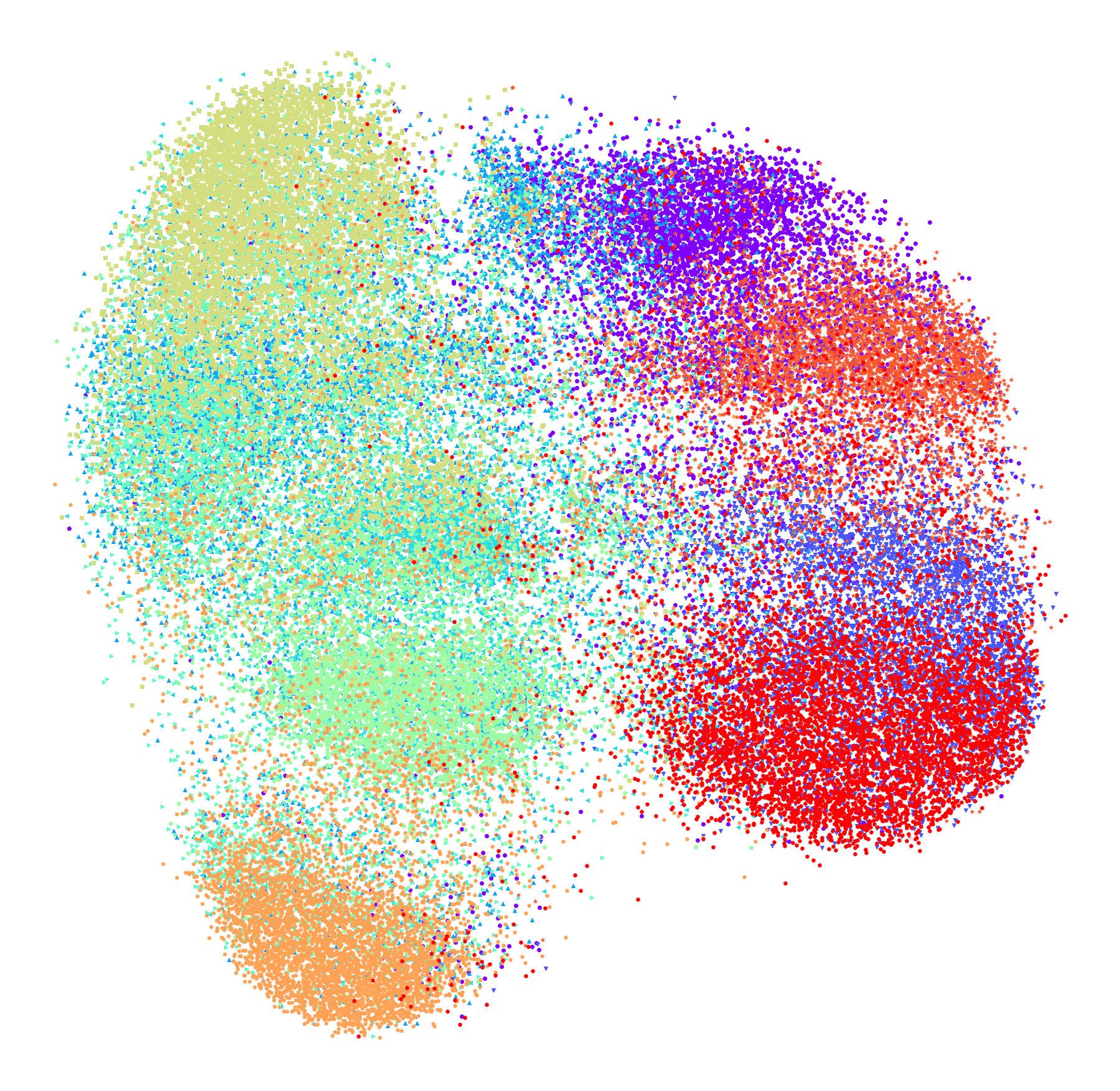}
        \caption{SIMI-ClusterGAN}
        \label{fig:iClusterGAN_cifar10}
    \end{subfigure}
    \caption{Encoding latent space visualisation for CIFAR-10 datset where colors are used to differentiate between distinct types of classes.}
    \label{fig:cifar10_encode_latent_space}
\end{figure}

\begin{figure}[ht!]
    \centering
    \begin{subfigure}[b]{0.22\textwidth}
        \includegraphics[width=\textwidth]{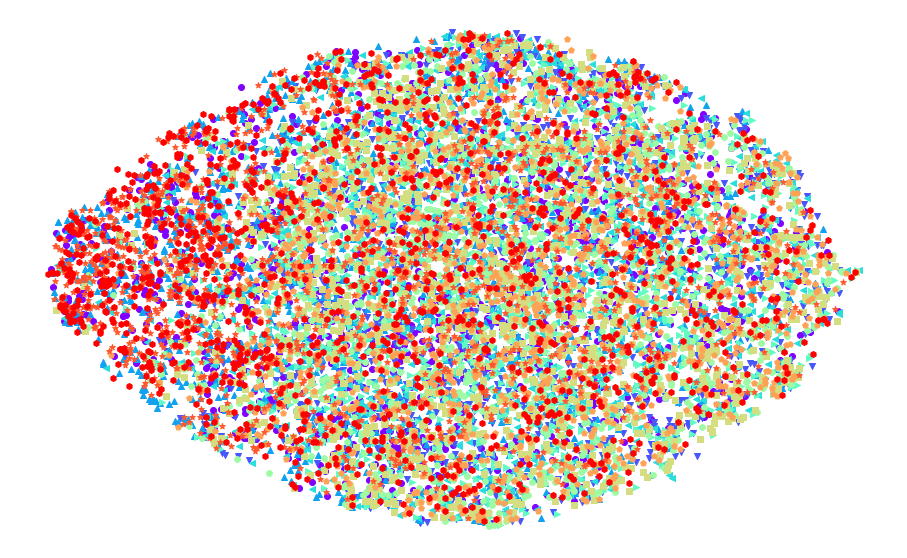}
        \caption{ClusterGAN}
        \label{fig:ClusterGAN_stl}
    \end{subfigure}
    ~ 
    \begin{subfigure}[b]{0.22\textwidth}
        \includegraphics[width=\textwidth]{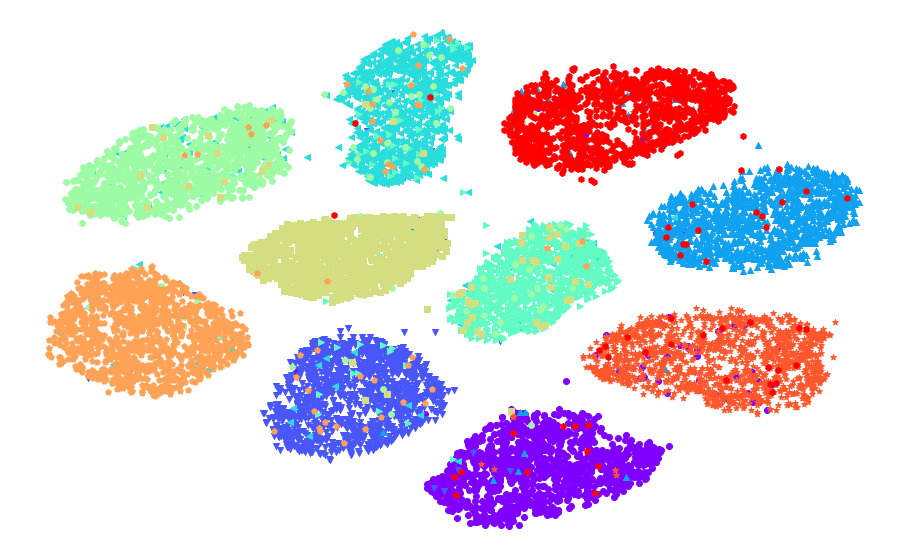}
        \caption{SIMI-ClusterGAN}
        \label{fig:iClusterGAN_stl}
    \end{subfigure}
    \caption{Encoding latent space visualisation for STL-10 datset where colors are used to differentiate between distinct types of classes.}
    \label{fig:stl_encode}
\end{figure}

When dealing with high-dimensional datasets such as CIFAR10 and STL-10, the disentanglement at $E_{\omega}$ is very important in determining the accuracy of clustering. In order to facilitate comprehension, the superiority of the SIMI-ClusterGAN over the ClusterGAN is displayed in Figure \ref{fig:cifar10_encode_latent_space}, which shows estimated encoded latent spaces from both methods in comparison to each other. Due to the joint learning framework of $E_{\omega}-G_{\theta}$ networks in the ClusterGAN method, $E_{\omega}$ is more reliable on generated sample semantics than focusing on downstream tasks like clustering. When we incorporate prior learning with three additional losses, the encoded latent spaces are well separated from ClusterGAN, which will lead to a significant improvement in clustering performance. Similar phenomena are also being observed for the STL-10 dataset. The encoded latent representation for the STL-10 dataset is depicted in Fig \ref{fig:stl_encode} where both clustering methods' results are shown.

\subsection{Latent space interpolation}
\begin{figure}[ht!]
    \centering
    \begin{subfigure}[b]{0.22\textwidth}
        \includegraphics[width=\textwidth]{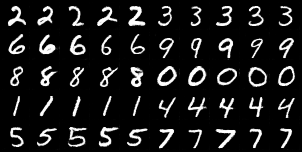}
        \caption{MNIST}
        \label{fig:mnist}
    \end{subfigure}
    ~ 
    \begin{subfigure}[b]{0.22\textwidth}
        \includegraphics[width=\textwidth]{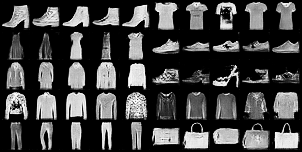}
        \caption{Fashion-MNIST}
        \label{fig:fashion-mnist}
    \end{subfigure}
    \caption{Latent space interpolation between classes}
    \label{fig:IClusterGAN_interploation}
\end{figure}

The one-hot code $(z_M)$ variable of the proposed SIMI-ClusterGAN method represents the cluster information in the data. The continuous variable ($z_n$) is controlling the variation of Gaussian representation in the latent space. In addition, we have also used the interpolation between different classes while varying two discrete variables $z_M^1$, $z_M^2$. The new continuous-discrete latent space vector is represented by $z=(z_n, \tau z_M^1 + (1-\tau)z_M^2)$ where $\tau \in[0,1]$, and $z_n$ is fixed. The smooth transition between one class to another class is depicted in Figure~\ref{fig:IClusterGAN_interploation} for the two datasets MNIST and Fashion-MNIST. 

\subsection{Ablation studies}
We have extensively analysed the proposed SIMI-ClusterGAN components in the following subsections.  

\subsubsection{different losses in SIMI-ClusterGAN}

\begin{figure}
    \centering
    \begin{subfigure}[b]{0.22\textwidth}
        \includegraphics[width=\textwidth]{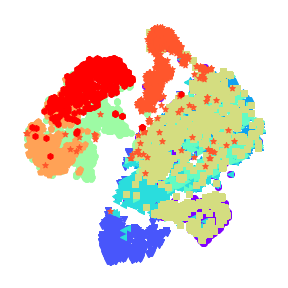}
        \caption{Raw-data}
        \label{fig:fashion_mnist_raw_tsne}
    \end{subfigure}
    \begin{subfigure}[b]{0.22\textwidth}
        \includegraphics[width=\textwidth]{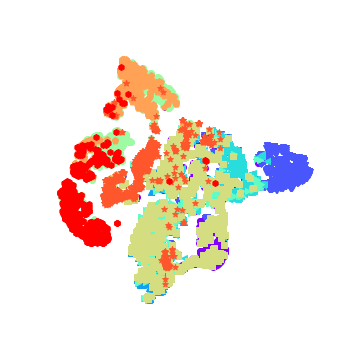}
        \caption{$J_{ce} + J_{mse} $}
        \label{fig:fashion_mnist_zn_zc_tsne}
    \end{subfigure}
    \begin{subfigure}[b]{0.22\textwidth}
        \includegraphics[width=\textwidth]{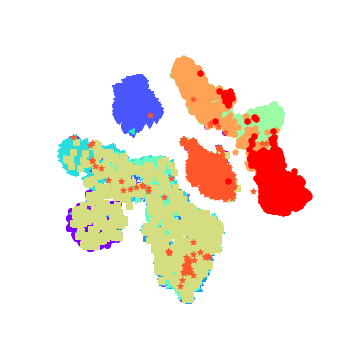}
        \caption{$J_{ce} + J_{mse} + J_{rec}$}
        \label{fig:fashion_mnist_zn_zc_recons_tsne}
    \end{subfigure}
   \begin{subfigure}[b]{0.22\textwidth}
        \includegraphics[width=\textwidth]{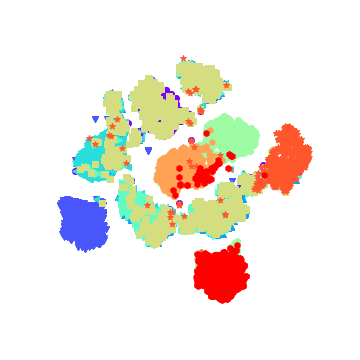}
        \caption{$J_{ce} + J_{mse} + J_{rec} +J_{cm} + J_{pce}$}
        \label{fig:fashion_mnist_all_losses_tsne}
    \end{subfigure}
    \caption{Visualisation of individual loss performance through t-SNE \cite{van2008visualizing}}
    \label{fig:Ablations studies}
\end{figure}

We also evaluate each loss function in the proposed SIMI-ClusterGAN objective function. The performance on Fashion-MNIST data by SIMI-ClusterGAN is reported in Table~\ref{Table:Dialation_studies}. In Table~\ref{Table:Dialation_studies}, the first loss is related when only the generated samples are passed to clustering inference network~\cite{mukherjee2019clustergan}. The uniform distribution based prior learning is not adequate to improve the clustering performance while using only generated samples. The performance is below the ClusterGAN performance. While considering the reconstruction loss into proposed objective function, the performance has improved significantly. All losses have an individual impact on the clustering performance to get accurate categories in fully unsupervised way. Figure~\ref{fig:Ablations studies} depicts  the encoding latent space in 2D for Fashion-MNIST dataset. Figure~\ref{fig:fashion_mnist_raw_tsne} depicts the raw-data in 2D encoding latent space. Considering the generated samples to the encoding network, the t-SNE \cite{van2008visualizing} plot of encoded latent space is illustrated in Figure~\ref{fig:fashion_mnist_zn_zc_tsne} which gives better representation than the raw-data. In addition, the encoding space gets better separated while considering the reconstruction loss into the proposed objective function. As illustrated in the Fig~\ref{fig:fashion_mnist_all_losses_tsne}, the SIMI-ClusterGAN gives better discrimination on encoding space while considering all the losses into proposed objective function. 

\begin{table}[]
\caption{ Effects of different losses on clustering performance of Fashion-MNIST dataset}
\centering
\begin{tabular}{llll}
\multirow{ 2}{*}{Dataset}                        & \multicolumn{2}{l}{Fashion-MNIST} \\ 
                                                & NMI       & ACC             \\ \hline
$J_{ce}$ + $J_{mse}$                                    & 0.5796    & 0.5853         \\ \hline
$J_{ce}$ + $J_{mse}$ + $J_{rec}$                      & 0.6580    & 0.6996        \\ \hline
 $J_{ce}$ + $J_{mse}$ + $J_{rec}$ +$J_{cm}$    & 0.6620    & 0.7056        \\ \hline
all losses                                      & 0.7053    & 0.7459      \\ \hline
\end{tabular}
\label{Table:Dialation_studies}
\end{table}

\subsubsection{Variations of $\sigma$ in $z_n$}
\begin{figure}[ht!]
    \centering
    \begin{subfigure}[b]{0.23\textwidth}
        \includegraphics[width=\textwidth]{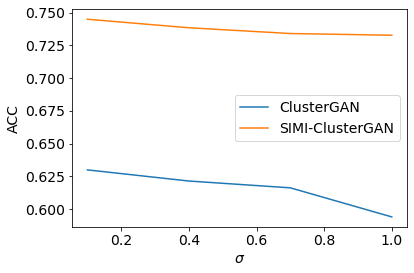}
        \caption{ACC}
        \label{fig:mnist_one_class_imbalanced}
    \end{subfigure}
    ~ 
    \begin{subfigure}[b]{0.23\textwidth}
        \includegraphics[width=\textwidth]{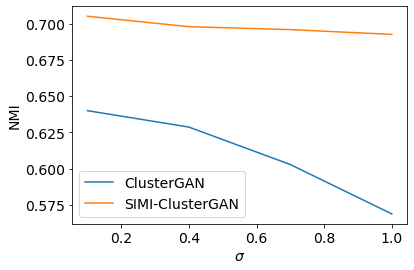}
        \caption{NMI}
        \label{fig:fashion-mnist}
    \end{subfigure}
    \caption{Varying $\sigma$ in $z_n$}
    \label{fig:varying_sigma}
\end{figure}

A non-smooth latent variable is formed when the continuous variable($z_n$) is mixed with the discrete variables($z_M$). Because the non-smooth bounded prior has an effect on clustering performance, it is thus necessary to assess the performance of clustering while changing the variance($\sigma$) of the continuous variable($z_n$). Therefore, we have illustrated the clustering performance on the Figure \ref{fig:varying_sigma} when $\sigma$ is varied incrementally by $0.3$. 
When the variance is at its highest ($\sigma =1.0$), ClusterGAN suffers a significant performance drop. Our suggested SIMI-ClusterGAN, in contrast to ClusterGAN, has no performance issues. Therefore, SIMI-ClusterGAN has a more robust performance than ClusterGAN.

\subsubsection{Uniform Prior based SIMI-ClusterGAN}
We also examined the clustering performance of our proposed method with the ClusterGAN when the $P_{\phi}$ network is not involved in the adversarial game. Thus, the SIMI-ClusterGAN prior comes from uniform distribution, as it is like ClusterGAN. Table \ref{tab:prior_uniform_comprasion} depicts the quantitative performance comparison between ClusterGAN and SIMI-ClusterGAN. It is clearly observed from the Table \ref{tab:prior_uniform_comprasion} that SIMI-ClusterGAN produces more disentanglement's by considering the real data distribution into $E_{\omega}$ network. However, for CIFAR-10 dataset, the adequate performance can't be obtained by considering uniform prior distribution. When the $E_{\omega}-G_{\theta}$ networks work together to learn, the $E_{\omega}$ focus is on semantics of the data rather than clustering. Therefore, prior learning is required for forcing $E_{\omega}$ network to match the real data distribution.

\begin{table}[ht!]
\caption{Uniform prior based SIMI-ClusterGAN performance}
\centering
\begin{tabular}{c|c|c|c|c}
\hline 
Methods & \multicolumn{2}{c|}{MNIST} & \multicolumn{2}{c}{CIFAR10}\tabularnewline
\hline 
 & ACC & NMI & ACC & NMI\tabularnewline
\hline 
ClusterGAN & 0.950 & 0.890 & 0.159 &0.025 \tabularnewline
\hline 
\textbf{SIMI-ClusterGAN} \textbackslash$P_{\phi}$  & \textbf{0.972} & \textbf{0.921} & \textbf{0.228} & \textbf{0.067}\tabularnewline
\hline 
\end{tabular}

\label{tab:prior_uniform_comprasion}
\end{table}



\subsection{Imbalanced Dataset Performance}

\begin{figure}
    \centering
    \begin{subfigure}[b]{0.22\textwidth}
        \includegraphics[width=\textwidth]{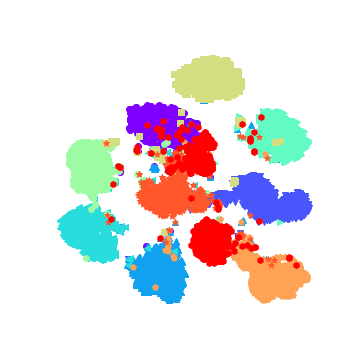}
        \caption{ClusterGAN}
        \label{fig:ClusterGAN_1_class}
    \end{subfigure}
    ~ 
    \begin{subfigure}[b]{0.22\textwidth}
        \includegraphics[width=\textwidth]{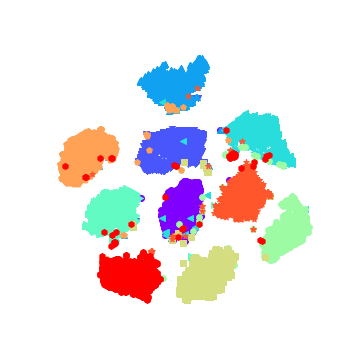}
        \caption{SIMI-ClusterGAN}
        \label{fig:iClusterGAN_1_class}
    \end{subfigure}
    \caption{Encoding latent space visualisation by t-SNE\cite{van2008visualizing} for one class imbalanced condition}
    \label{fig:1_class_imbalanced_tsne_pca_plots}
\end{figure}

We have also validated the performance of the proposed method with ClusterGAN under two different imbalanced conditions: single class imbalance and multi-class imbalance. MNIST dataset has been considered in which two class imbalanced settings have been tested. For single class imbalanced settings, $0.1$ fraction of zero samples (\textquoteleft digit-0\textquoteright) and remaining all the samples have been taken for training whereas all the test samples($10K$) are used for testing the clustering performance. Similarly for multi-class imbalanced settings, three imbalanced classes are \textquoteleft digit-1\textquoteright, \textquoteleft digit-3\textquoteright and \textquoteleft digit-5\textquoteright with fraction of  $0.1,0.3\ $,  and  $\, 0.5$. Similar to the one class-imbalanced case, all the testing data samples are used to test the clustering inference network. The performance of the proposed SIMI-ClusterGAN is shown in the Table~\ref{table:Imbalanced_clustering_performance}. It is clearly observed from the Table~\ref{table:Imbalanced_clustering_performance} that our proposed method  out-performs uniform prior based ClusterGAN performance. For better understanding of $E_{\omega}$ performance for both the methods, we have visualised encoding latent space is shown in  Figure~\ref{fig:1_class_imbalanced_tsne_pca_plots}. It is clearly observed from the figure that our proposed method gives better separable decision boundary than the ClusterGAN.  

\begin{table}[ht!]
\centering
\caption{Imbalanced Clustering Performance}
\begin{tabular}{c|c|c|c|c}
\hline 
Methods & \multicolumn{4}{c}{MNIST}\tabularnewline
\hline 
 & \multicolumn{2}{c|}{1-Class} & \multicolumn{2}{c}{3-Classes}\tabularnewline
\hline 
 & ACC & NMI & ACC & NMI\tabularnewline
\hline 
ClusterGAN & 0.892 & 0.836 & 0.812 & 0.787\tabularnewline
\hline 
\textbf{SIMI-ClusterGAN} & \textbf{0.969}& \textbf{0.923} & \textbf{0.949} & \textbf{0.909}\tabularnewline
\hline 
\end{tabular}
\label{table:Imbalanced_clustering_performance}
\end{table}

\begin{figure}[ht!]
    \centering
    \begin{subfigure}[b]{0.23\textwidth}
        \includegraphics[width=\textwidth]{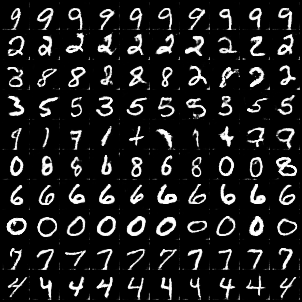}
        \caption{ClusterGAN}
        \label{fig:mnist_one_class_imbalanced}
    \end{subfigure}
    ~ 
    \begin{subfigure}[b]{0.23\textwidth}
        \includegraphics[width=\textwidth]{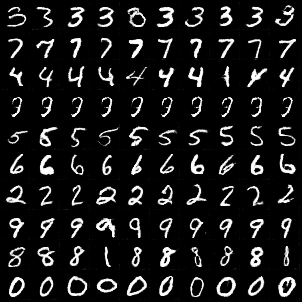}
        \caption{SIMI-ClusterGAN}
        \label{fig:fashion-mnist}
    \end{subfigure}
    \caption{Samples generated by ClusterGAN and SIMI-ClusterGAN under three classes imbalanced condition.}
    \label{fig:3_class_imbalanced}
\end{figure}

\begin{figure}[ht!]
    \centering
    \begin{subfigure}[b]{0.23\textwidth}
        \includegraphics[width=\textwidth]{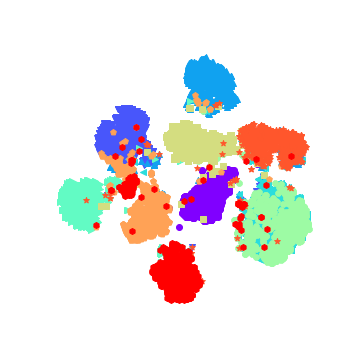}
        \caption{ClusterGAN}
        \label{fig:ClusterGAN_three_class_imbalanced}
    \end{subfigure}
    ~ 
    \begin{subfigure}[b]{0.23\textwidth}
        \includegraphics[width=\textwidth]{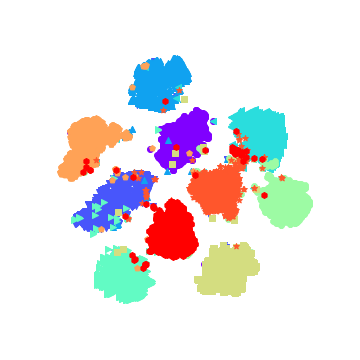}
        \caption{SIMI-ClusterGAN}
        \label{fig:SIMI-ClusterGAN_three_class_imbalanced}
    \end{subfigure}
    \caption{Encoding latent space visualisation by t-SNE\cite{van2008visualizing} for three classes imbalanced condition}
    \label{fig:3_class_imbalanced_tsne_pca_plots}
\end{figure}

\begin{figure}[ht!]
    \centering
    \begin{subfigure}[b]{0.23\textwidth}
        \includegraphics[width=\textwidth]{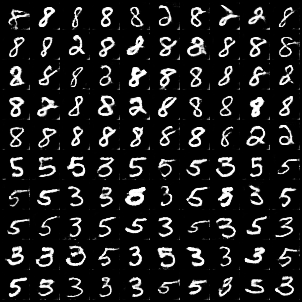}
        \caption{ClusterGAN:'digit-3' and 'digit-5'}
        \label{fig:ClusterGAN_three_class_imbalanced}
    \end{subfigure}
    ~ 
    \begin{subfigure}[b]{0.23\textwidth}
        \includegraphics[width=\textwidth]{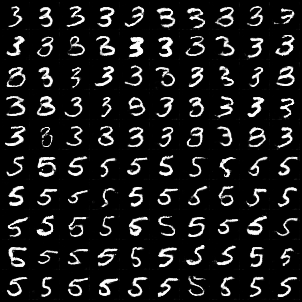}
        \caption{SIMI-ClusterGAN: 'digit-3' and 'digit-5'}
        \label{fig:SIMI-ClusterGAN_three_class_imbalanced}
    \end{subfigure}
    \caption{Two minor classes generated by ClusterGAN and SIMI-ClusterGAN}
    \label{fig:2_minor_class_generation}
\end{figure}

For three classes imbalanced conditions, both the methods failed to generate the minor \textquoteleft digit-1\textquoteright which is shown in the Figure~\ref{fig:3_class_imbalanced}. However, for other two minor classes (\textquoteleft digit-3\textquoteright and \textquoteleft digit-5\textquoteright), proposed SIMI-ClusterGAN is able to generate pure classes. For better visual understanding, we have plotted with two minor classes data in Figure~\ref{fig:2_minor_class_generation} in which \textquoteleft 50\textquoteright samples were generated for each class. It is clearly observed from the Figure~\ref{fig:2_minor_class_generation} that our proposed method is able to generate both classes with high purity. However, ClusterGAN, generated mixture of samples of \textquoteleft digit-3\textquoteright and \textquoteleft digit-5\textquoteright and $50$ samples were generated for digit-8. We have also visualised encoding latent space in 2D for both the methods in Figure~\ref{fig:3_class_imbalanced_tsne_pca_plots}. 

\squeezeup

\section{conclusion}

In this paper, we have proposed a SIMI-ClusterGAN which is the modification of uniform prior latent space based ClusterGAN method. The proposed method consists of four networks, a prior network, a generator, a discriminator and a clustering inference network. The proposed method is followed by two stage learning methods: learning categorical prior from the data directly and then the learned prior is used to modify clustering performance through adversarial training between conditional generator and discriminator. The simulation results indicate that the proposed SIMI-ClusterGAN achieved better performance than the state-of-the-art methods. The effectiveness of each loss has been studied through an ablation study. It is also observed that the proposed method can handle imbalanced conditions. However, if the data distribution is highly imbalanced in nature, the SIM-based $P_{\phi}$ won't be able to capture how the real data is distributed.    Consequently, SIMI-ClusterGAN fails to generate the minority classes. Rather than relying on two-stage techniques, we will instead leverage self-supervised losses\cite{li2021contrastive} to develop a single-stage learning technique that simultaneously updates prior and clustering inference networks.

\bibliographystyle{IEEEtran}

\bibliography{reference}

\begin{thebibliography}{10}
\providecommand{\url}[1]{#1}
\csname url@samestyle\endcsname
\providecommand{\newblock}{\relax}
\providecommand{\bibinfo}[2]{#2}
\providecommand{\BIBentrySTDinterwordspacing}{\spaceskip=0pt\relax}
\providecommand{\BIBentryALTinterwordstretchfactor}{4}
\providecommand{\BIBentryALTinterwordspacing}{\spaceskip=\fontdimen2\font plus
\BIBentryALTinterwordstretchfactor\fontdimen3\font minus
  \fontdimen4\font\relax}
\providecommand{\BIBforeignlanguage}[2]{{%
\expandafter\ifx\csname l@#1\endcsname\relax
\typeout{** WARNING: IEEEtran.bst: No hyphenation pattern has been}%
\typeout{** loaded for the language `#1'. Using the pattern for}%
\typeout{** the default language instead.}%
\else
\language=\csname l@#1\endcsname
\fi
#2}}
\providecommand{\BIBdecl}{\relax}
\BIBdecl

\bibitem{chuang2006fuzzy}
K.-S. Chuang, H.-L. Tzeng, S.~Chen, J.~Wu, and T.-J. Chen, ``Fuzzy c-means
  clustering with spatial information for image segmentation,''
  \emph{computerized medical imaging and graphics}, vol.~30, no.~1, pp. 9--15,
  2006.

\bibitem{caron2018deep}
M.~Caron, P.~Bojanowski, A.~Joulin, and M.~Douze, ``Deep clustering for
  unsupervised learning of visual features,'' in \emph{Proceedings of the
  European Conference on Computer Vision (ECCV)}, 2018, pp. 132--149.

\bibitem{vincent2010stacked}
P.~Vincent, H.~Larochelle, I.~Lajoie, Y.~Bengio, P.-A. Manzagol, and L.~Bottou,
  ``Stacked denoising autoencoders: Learning useful representations in a deep
  network with a local denoising criterion.'' \emph{Journal of machine learning
  research}, vol.~11, no.~12, 2010.

\bibitem{yang2017towards}
B.~Yang, X.~Fu, N.~D. Sidiropoulos, and M.~Hong, ``Towards k-means-friendly
  spaces: Simultaneous deep learning and clustering,'' in \emph{international
  conference on machine learning}, 2017, pp. 3861--3870.

\bibitem{xie2016unsupervised}
J.~Xie, R.~Girshick, and A.~Farhadi, ``Unsupervised deep embedding for
  clustering analysis,'' in \emph{International conference on machine
  learning}, 2016, pp. 478--487.

\bibitem{yang2019deep}
X.~Yang, C.~Deng, F.~Zheng, J.~Yan, and W.~Liu, ``Deep spectral clustering
  using dual autoencoder network,'' in \emph{Proceedings of the IEEE Conference
  on Computer Vision and Pattern Recognition}, 2019, pp. 4066--4075.

\bibitem{mukherjee2019clustergan}
S.~Mukherjee, H.~Asnani, E.~Lin, and S.~Kannan, ``Clustergan: Latent space
  clustering in generative adversarial networks,'' in \emph{Proceedings of the
  AAAI Conference on Artificial Intelligence}, vol.~33, 2019, pp. 4610--4617.

\bibitem{min2018survey}
E.~Min, X.~Guo, Q.~Liu, G.~Zhang, J.~Cui, and J.~Long, ``A survey of clustering
  with deep learning: From the perspective of network architecture,''
  \emph{IEEE Access}, vol.~6, pp. 39\,501--39\,514, 2018.

\bibitem{macqueen1967some}
J.~MacQueen \emph{et~al.}, ``Some methods for classification and analysis of
  multivariate observations,'' in \emph{Proceedings of the fifth Berkeley
  symposium on mathematical statistics and probability}, vol.~1, no.~14.\hskip
  1em plus 0.5em minus 0.4em\relax Oakland, CA, USA, 1967, pp. 281--297.

\bibitem{goodfellow2014generative}
I.~Goodfellow, J.~Pouget-Abadie, M.~Mirza, B.~Xu, D.~Warde-Farley, S.~Ozair,
  A.~Courville, and Y.~Bengio, ``Generative adversarial nets,'' in
  \emph{Advances in neural information processing systems}, 2014, pp.
  2672--2680.

\bibitem{ma2020snegan}
L.~Ma, Y.~Ma, Q.~Lin, J.~Ji, C.~A.~C. Coello, and M.~Gong, ``Snegan: Signed
  network embedding by using generative adversarial nets,'' \emph{IEEE
  Transactions on Emerging Topics in Computational Intelligence}, 2020.

\bibitem{he2021finger}
J.~He, L.~Shen, Y.~Yao, H.~Wang, G.~Zhao, X.~Gu, and W.~Ding, ``Finger vein
  image deblurring using neighbors-based binary-gan (nb-gan),'' \emph{IEEE
  Transactions on Emerging Topics in Computational Intelligence}, 2021.

\bibitem{li2019af}
Q.~Li, H.~Qu, Z.~Liu, N.~Zhou, W.~Sun, S.~Sigg, and J.~Li, ``Af-dcgan:
  Amplitude feature deep convolutional gan for fingerprint construction in
  indoor localization systems,'' \emph{IEEE Transactions on Emerging Topics in
  Computational Intelligence}, 2019.

\bibitem{isola2017image}
P.~Isola, J.-Y. Zhu, T.~Zhou, and A.~A. Efros, ``Image-to-image translation
  with conditional adversarial networks,'' in \emph{Proceedings of the IEEE
  conference on computer vision and pattern recognition}, 2017, pp. 1125--1134.

\bibitem{oord2016wavenet}
A.~v.~d. Oord, S.~Dieleman, H.~Zen, K.~Simonyan, O.~Vinyals, A.~Graves,
  N.~Kalchbrenner, A.~Senior, and K.~Kavukcuoglu, ``Wavenet: A generative model
  for raw audio,'' \emph{arXiv preprint arXiv:1609.03499}, 2016.

\bibitem{tang2020multi}
H.~Tang, D.~Xu, Y.~Yan, J.~J. Corso, P.~H. Torr, and N.~Sebe, ``Multi-channel
  attention selection gans for guided image-to-image translation,'' \emph{arXiv
  preprint arXiv:2002.01048}, 2020.

\bibitem{jing2020self}
L.~Jing and Y.~Tian, ``Self-supervised visual feature learning with deep neural
  networks: A survey,'' \emph{IEEE Transactions on Pattern Analysis and Machine
  Intelligence}, 2020.

\bibitem{liu2020cgan}
C.-H. Liu, H.~Chang, and T.~Park, ``Da-cgan: A framework for indoor radio
  design using a dimension-aware conditional generative adversarial network,''
  in \emph{Proceedings of the IEEE/CVF Conference on Computer Vision and
  Pattern Recognition Workshops}, 2020, pp. 498--499.

\bibitem{zheng2021reward}
C.~Zheng, S.~Yang, J.~M. Parra-Ullauri, A.~Garcia-Dominguez, and N.~Bencomo,
  ``Reward-reinforced generative adversarial networks for multi-agent
  systems,'' \emph{IEEE Transactions on Emerging Topics in Computational
  Intelligence}, 2021.

\bibitem{chen2016infogan}
X.~Chen, Y.~Duan, R.~Houthooft, J.~Schulman, I.~Sutskever, and P.~Abbeel,
  ``Infogan: Interpretable representation learning by information maximizing
  generative adversarial nets,'' in \emph{Advances in neural information
  processing systems}, 2016, pp. 2172--2180.

\bibitem{arora2018gans}
S.~Arora, A.~Risteski, and Y.~Zhang, ``Do gans learn the distribution? some
  theory and empirics,'' in \emph{International Conference on Learning
  Representations}, 2018.

\bibitem{karras2018progressive}
T.~Karras, T.~Aila, S.~Laine, and J.~Lehtinen, ``Progressive growing of gans
  for improved quality, stability, and variation,'' in \emph{International
  Conference on Learning Representations}, 2018.

\bibitem{mishra2020effect}
D.~Mishra, J.~Aravind, and A.~Prathosh, ``Effect of the latent structure on
  clustering with gans,'' \emph{IEEE Signal Processing Letters}, 2020.

\bibitem{chang2017deep}
J.~Chang, L.~Wang, G.~Meng, S.~Xiang, and C.~Pan, ``Deep adaptive image
  clustering,'' in \emph{Proceedings of the IEEE international conference on
  computer vision}, 2017, pp. 5879--5887.

\bibitem{yang2016joint}
J.~Yang, D.~Parikh, and D.~Batra, ``Joint unsupervised learning of deep
  representations and image clusters,'' in \emph{Proceedings of the IEEE
  Conference on Computer Vision and Pattern Recognition}, 2016, pp. 5147--5156.

\bibitem{guo2017improved}
X.~Guo, L.~Gao, X.~Liu, and J.~Yin, ``Improved deep embedded clustering with
  local structure preservation.'' in \emph{IJCAI}, 2017, pp. 1753--1759.

\bibitem{mathieu2016disentangling}
M.~F. Mathieu, J.~J. Zhao, J.~Zhao, A.~Ramesh, P.~Sprechmann, and Y.~LeCun,
  ``Disentangling factors of variation in deep representation using adversarial
  training,'' in \emph{Advances in neural information processing systems},
  2016, pp. 5040--5048.

\bibitem{gonzalez2018image}
A.~Gonzalez-Garcia, J.~Van De~Weijer, and Y.~Bengio, ``Image-to-image
  translation for cross-domain disentanglement,'' in \emph{Advances in neural
  information processing systems}, 2018, pp. 1287--1298.

\bibitem{tschannen2018recent}
M.~Tschannen, O.~Bachem, and M.~Lucic, ``Recent advances in autoencoder-based
  representation learning,'' \emph{arXiv preprint arXiv:1812.05069}, 2018.

\bibitem{pan2020loss}
Z.~Pan, W.~Yu, B.~Wang, H.~Xie, V.~S. Sheng, J.~Lei, and S.~Kwong, ``Loss
  functions of generative adversarial networks (gans): Opportunities and
  challenges,'' \emph{IEEE Transactions on Emerging Topics in Computational
  Intelligence}, vol.~4, no.~4, pp. 500--522, 2020.

\bibitem{hadad2018two}
N.~Hadad, L.~Wolf, and M.~Shahar, ``A two-step disentanglement method,'' in
  \emph{Proceedings of the IEEE Conference on Computer Vision and Pattern
  Recognition}, 2018, pp. 772--780.

\bibitem{patacchiola2019autoencoders}
M.~Patacchiola, P.~Fox-Roberts, and E.~Rosten, ``Y-autoencoders: disentangling
  latent representations via sequential-encoding,'' \emph{arXiv preprint
  arXiv:1907.10949}, 2019.

\bibitem{ye2020probabilistic}
M.~Ye and J.~Shen, ``Probabilistic structural latent representation for
  unsupervised embedding,'' in \emph{Proceedings of the IEEE/CVF Conference on
  Computer Vision and Pattern Recognition}, 2020, pp. 5457--5466.

\bibitem{chen2018isolating}
R.~T. Chen, X.~Li, R.~B. Grosse, and D.~K. Duvenaud, ``Isolating sources of
  disentanglement in variational autoencoders,'' in \emph{Advances in Neural
  Information Processing Systems}, 2018, pp. 2610--2620.

\bibitem{higgins2016beta}
I.~Higgins, L.~Matthey, A.~Pal, C.~Burgess, X.~Glorot, M.~Botvinick,
  S.~Mohamed, and A.~Lerchner, ``beta-vae: Learning basic visual concepts with
  a constrained variational framework,'' 2016.

\bibitem{kim2018disentangling}
H.~Kim and A.~Mnih, ``Disentangling by factorising,'' \emph{arXiv preprint
  arXiv:1802.05983}, 2018.

\bibitem{dupont2018learning}
E.~Dupont, ``Learning disentangled joint continuous and discrete
  representations,'' in \emph{Advances in Neural Information Processing
  Systems}, 2018, pp. 710--720.

\bibitem{gurumurthy2017deligan}
S.~Gurumurthy, R.~Kiran~Sarvadevabhatla, and R.~Venkatesh~Babu, ``Deligan:
  Generative adversarial networks for diverse and limited data,'' in
  \emph{Proceedings of the IEEE conference on computer vision and pattern
  recognition}, 2017, pp. 166--174.

\bibitem{gulrajani2017improved}
I.~Gulrajani, F.~Ahmed, M.~Arjovsky, V.~Dumoulin, and A.~C. Courville,
  ``Improved training of wasserstein gans,'' in \emph{Advances in neural
  information processing systems}, 2017, pp. 5767--5777.

\bibitem{dam2021does}
T.~Dam, M.~M. Ferdaus, S.~G. Anavatti, S.~Jayavelu, and H.~A. Abbass, ``Does
  adversarial oversampling help us?'' in \emph{Proceedings of the 30th ACM
  International Conference on Information \& Knowledge Management}, 2021, pp.
  2970--2973.

\bibitem{creswell2018inverting}
A.~Creswell and A.~A. Bharath, ``Inverting the generator of a generative
  adversarial network,'' \emph{IEEE transactions on neural networks and
  learning systems}, vol.~30, no.~7, pp. 1967--1974, 2018.

\bibitem{lipton2017precise}
Z.~C. Lipton and S.~Tripathi, ``Precise recovery of latent vectors from
  generative adversarial networks,'' \emph{arXiv preprint arXiv:1702.04782},
  2017.

\bibitem{krause2010discriminative}
A.~Krause, P.~Perona, and R.~G. Gomes, ``Discriminative clustering by
  regularized information maximization,'' in \emph{Advances in neural
  information processing systems}, 2010, pp. 775--783.

\bibitem{hu2017learning}
W.~Hu, T.~Miyato, S.~Tokui, E.~Matsumoto, and M.~Sugiyama, ``Learning discrete
  representations via information maximizing self-augmented training,''
  \emph{arXiv preprint arXiv:1702.08720}, 2017.

\bibitem{grandvalet2005semi}
Y.~Grandvalet and Y.~Bengio, ``Semi-supervised learning by entropy
  minimization,'' in \emph{Advances in neural information processing systems},
  2005, pp. 529--536.

\bibitem{miyato2015distributional}
T.~Miyato, S.-i. Maeda, M.~Koyama, K.~Nakae, and S.~Ishii, ``Distributional
  smoothing with virtual adversarial training,'' \emph{arXiv preprint
  arXiv:1507.00677}, 2015.

\bibitem{zhao2018adversarially}
J.~Zhao, Y.~Kim, K.~Zhang, A.~Rush, and Y.~LeCun, ``Adversarially regularized
  autoencoders,'' in \emph{International Conference on Machine Learning}.\hskip
  1em plus 0.5em minus 0.4em\relax PMLR, 2018, pp. 5902--5911.

\bibitem{mao2019mode}
Q.~Mao, H.-Y. Lee, H.-Y. Tseng, S.~Ma, and M.-H. Yang, ``Mode seeking
  generative adversarial networks for diverse image synthesis,'' in
  \emph{Proceedings of the IEEE Conference on Computer Vision and Pattern
  Recognition}, 2019, pp. 1429--1437.

\bibitem{coates2011analysis}
A.~Coates, A.~Ng, and H.~Lee, ``An analysis of single-layer networks in
  unsupervised feature learning,'' in \emph{Proceedings of the fourteenth
  international conference on artificial intelligence and statistics}, 2011,
  pp. 215--223.

\bibitem{krizhevsky2009learning}
A.~Krizhevsky, G.~Hinton \emph{et~al.}, ``Learning multiple layers of features
  from tiny images,'' 2009.

\bibitem{xiao2017fashion}
H.~Xiao, K.~Rasul, and R.~Vollgraf, ``Fashion-mnist: a novel image dataset for
  benchmarking machine learning algorithms,'' \emph{arXiv preprint
  arXiv:1708.07747}, 2017.

\bibitem{lee1999learning}
D.~D. Lee and H.~S. Seung, ``Learning the parts of objects by non-negative
  matrix factorization,'' \emph{Nature}, vol. 401, no. 6755, pp. 788--791,
  1999.

\bibitem{shi2000normalized}
J.~Shi and J.~Malik, ``Normalized cuts and image segmentation,'' \emph{IEEE
  Transactions on pattern analysis and machine intelligence}, vol.~22, no.~8,
  pp. 888--905, 2000.

\bibitem{zhang2012graph}
W.~Zhang, X.~Wang, D.~Zhao, and X.~Tang, ``Graph degree linkage: Agglomerative
  clustering on a directed graph,'' in \emph{European Conference on Computer
  Vision}.\hskip 1em plus 0.5em minus 0.4em\relax Springer, 2012, pp. 428--441.

\bibitem{ghasedi2017deep}
K.~Ghasedi~Dizaji, A.~Herandi, C.~Deng, W.~Cai, and H.~Huang, ``Deep clustering
  via joint convolutional autoencoder embedding and relative entropy
  minimization,'' in \emph{Proceedings of the IEEE international conference on
  computer vision}, 2017, pp. 5736--5745.

\bibitem{shaham2018spectralnet}
U.~Shaham, K.~Stanton, H.~Li, B.~Nadler, R.~Basri, and Y.~Kluger,
  ``Spectralnet: Spectral clustering using deep neural networks,'' \emph{arXiv
  preprint arXiv:1801.01587}, 2018.

\bibitem{springenberg2015unsupervised}
J.~T. Springenberg, ``Unsupervised and semi-supervised learning with
  categorical generative adversarial networks,'' \emph{arXiv preprint
  arXiv:1511.06390}, 2015.

\bibitem{yu2018mixture}
Y.~Yu and W.-J. Zhou, ``Mixture of gans for clustering.'' in \emph{IJCAI},
  2018, pp. 3047--3053.

\bibitem{kuhn1955hungarian}
H.~W. Kuhn, ``The hungarian method for the assignment problem,'' \emph{Naval
  research logistics quarterly}, vol.~2, no. 1-2, pp. 83--97, 1955.

\bibitem{he2016deep}
K.~He, X.~Zhang, S.~Ren, and J.~Sun, ``Deep residual learning for image
  recognition,'' in \emph{Proceedings of the IEEE conference on computer vision
  and pattern recognition}, 2016, pp. 770--778.

\bibitem{jiang2017variational}
Z.~Jiang, Y.~Zheng, H.~Tan, B.~Tang, and H.~Zhou, ``Variational deep embedding:
  an unsupervised and generative approach to clustering,'' in \emph{Proceedings
  of the 26th International Joint Conference on Artificial Intelligence}, 2017,
  pp. 1965--1972.

\bibitem{sarfraz2019efficient}
S.~Sarfraz, V.~Sharma, and R.~Stiefelhagen, ``Efficient parameter-free
  clustering using first neighbor relations,'' in \emph{Proceedings of the
  IEEE/CVF Conference on Computer Vision and Pattern Recognition}, 2019, pp.
  8934--8943.

\bibitem{van2008visualizing}
L.~Van~der Maaten and G.~Hinton, ``Visualizing data using t-sne.''
  \emph{Journal of machine learning research}, vol.~9, no.~11, 2008.

\bibitem{li2021contrastive}
Y.~Li, P.~Hu, Z.~Liu, D.~Peng, J.~T. Zhou, and X.~Peng, ``Contrastive
  clustering,'' in \emph{2021 AAAI Conference on Artificial Intelligence
  (AAAI)}, 2021.

\end{thebibliography}

\end{document}